\documentclass[10pt, a4paper]{article}

\usepackage{lrec-coling2024} 

\usepackage{xcolor}
\usepackage{hyperref}
 \definecolor{darkblue}{rgb}{0, 0, 0.5}
  \hypersetup{colorlinks=true, citecolor=darkblue, linkcolor=darkblue, urlcolor=darkblue}

\definecolor{overlap_full}{HTML}{ffc9c9}
\definecolor{overlap_lexi}{HTML}{a5d8ff}
\definecolor{doclength_full}{HTML}{ffe8cc}
\definecolor{doclength_rst}{HTML}{99e9f2}

\usepackage{xstring}

\usepackage{color}

\usepackage{booktabs}
\usepackage{makecell}
\usepackage{pifont} 
\usepackage{multirow}
\usepackage{array}
\usepackage{graphicx}
\usepackage{subcaption}
\usepackage{amsfonts}
\usepackage{amsmath}
\usepackage{bm}
\usepackage{pgfplots}

\usepackage{CJKutf8}

\newcommand{\cmark}{\ding{51}}%
\newcommand{\xmark}{\ding{55}}%

\definecolor{greenrgb}{rgb}{0.18, 0.71, 0.18}
\definecolor{lightblue1}{rgb}{0.6, 0.81, 0.93}
\definecolor{lightblue2}{rgb}{0.45, 0.76, 0.98}
\definecolor{lightblue3}{rgb}{0.35, 0.76, 0.98}
\definecolor{myblue}{HTML}{0070C0}

\definecolor{lightwhite}{HTML}{B7EAEB}

\newcommand{\paratitle}[1]{\vspace{1.3ex}\noindent \textbf{#1}}

\title{\vspace*{.5\baselineskip} \textbf{Enhancing Cross-Document Event Coreference Resolution by Discourse Structure and Semantic Information}}

\name{
\begin{tabular}{c}
  Qiang Gao$^{1}$, Bobo Li$^{1}$, Zixiang Meng$^{1}$, Yunlong Li$^{1}$, 
  Jun Zhou$^{1}$, Fei Li$^{1}$, \\
 Chong Teng$^{1}$, Donghong Ji$^{1}$
\end{tabular}
} 

\address{
$^{1}$Key Laboratory of Aerospace Information Security and Trusted Computing, Ministry of Education, \\School of Cyber Science and Engineering, Wuhan University \\
 gaoqiang.nlp@gmail.com, \\ \{boboli, zixiangmeng, yunlongli, j$.$zhou, lifei\_csnlp, tengchong, dhji\}@whu.edu.cn\\}

\abstract{
Existing cross-document event coreference resolution models, 
which either compute mention similarity directly or enhance mention representation by extracting event arguments (such as location, time, agent, and patient), 
lacking the ability to utilize document-level information. As a result, 
they struggle to capture long-distance dependencies. 
This shortcoming leads to their underwhelming performance in determining coreference for 
the events where their argument information relies on long-distance dependencies. 
In light of these limitations,  we propose the construction of document-level Rhetorical Structure Theory (RST) trees and cross-document Lexical Chains to model the structural and semantic information of documents.
Subsequently, cross-document heterogeneous graphs are constructed and GAT is utilized to learn the representations of events.
Finally, a pair scorer calculates the similarity between each pair of events and co-referred events can be recognized using standard clustering algorithm.
Additionally, as the existing cross-document event coreference datasets are limited to English, we have developed a large-scale Chinese cross-document event coreference dataset to fill this gap, which comprises 53,066 event mentions and 4,476 clusters.
After applying our model on the English and Chinese datasets respectively, 
it outperforms all baselines by large margins.
 \\ \newline  }

\begin{document}

\maketitleabstract

\section{Introduction}

\begin{figure*}[!ht]
\centering
    \begin{subfigure}{\textwidth}
        \includegraphics[width=\textwidth]{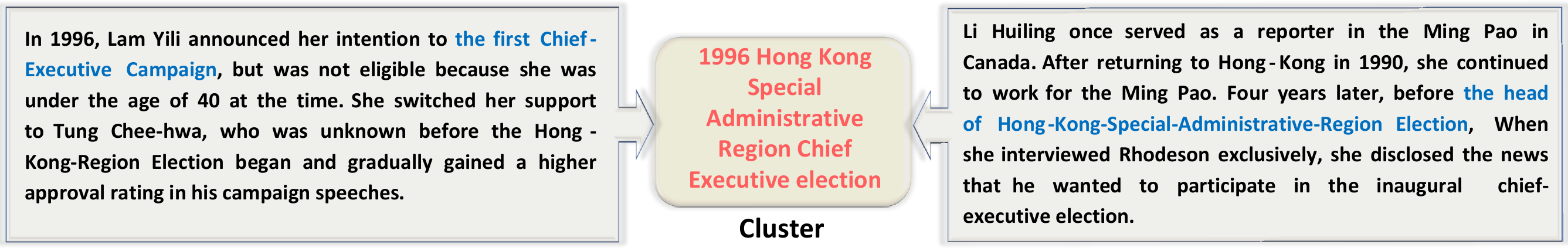}
        \caption{A sample illustration of two co-referential events.}
        \label{fig1.a}
    \end{subfigure}
    \hfill
    \begin{subfigure}{\textwidth}
        \includegraphics[width=\textwidth]{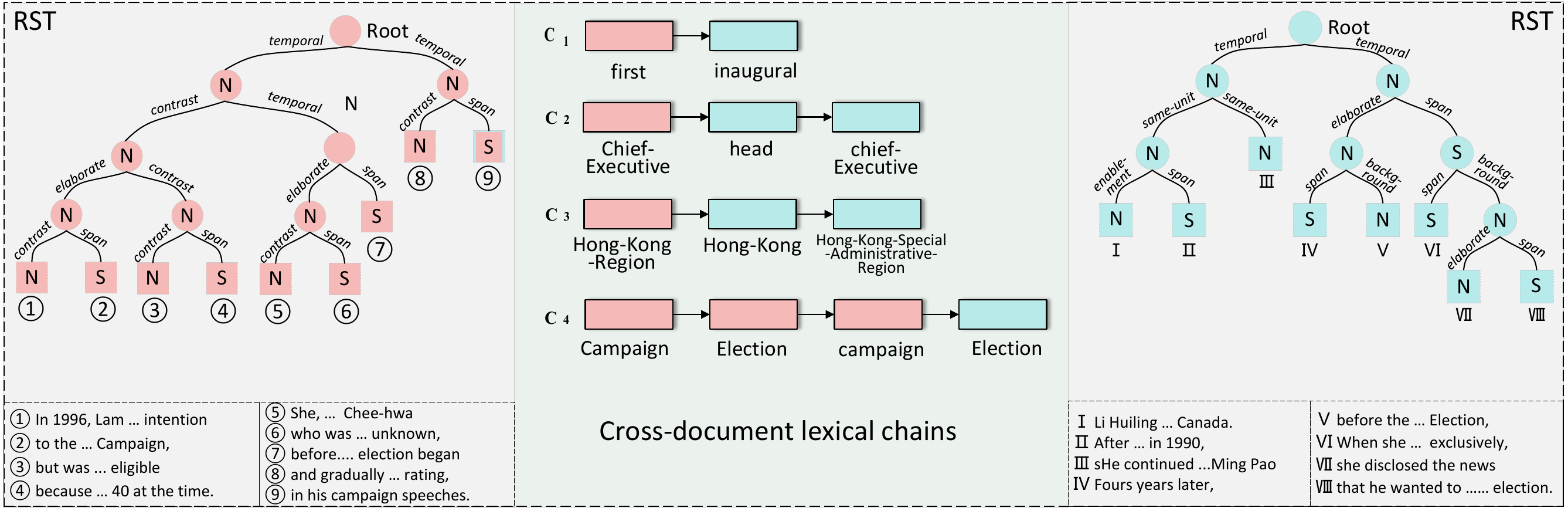}
        \caption{Discourse information: document-level RST trees and cross-document lexical chains.}
        \label{fig1.b}
    \end{subfigure}

\caption{An example to show cross-document event coreference resolution and our main idea of building
document and cross-document information.
\textcolor{myblue}{the first Chief-Executive election} and \textcolor{myblue}{the head of Hong Kong Special Administrative Region Election} refer to the same event ``1996 Hong Kong Special Administrative Region Chief Executive election". 
}
\label{fig.1}
\end{figure*}

Cross-document event coreference resolution (CDECR) is a critical task in natural language processing (NLP), exhibiting substantial applicability across various downstream tasks, including information extraction \citep{resin}, document summarization \citep{leveraging}, question-answering \citep{baleen}, etc.
As shown in Figure~\ref{fig1.a}, the primary goal of CDECR is to identify text references across multiple documents that relate to the same event.
Given the necessity to comprehend an array of documents concurrently, CDECR is more challenging than within-document event coreference resolution \citep{word-level}.

Recognizing these complexities, several methods have been proposed to address the challenges inherent to CDECR.
\citet{CD2CR} utilized event mention representation to compute coreference scores, 
It often misjudges mentions that appear similar.
\citet{pairwise} proposed extracting event arguments to enhance the mention representation, 
but this approach is heavily reliant on the performance of the SRL (semantic role labeling) tool, 
unable to model discourse-level information.
The most commonly used CDECR dataset ECB+ \citep{ecb+}, suffers from the limitations in terms of small data scale and event categorization, as it only addresses coreference within the events with the same type, which does not align well with real-world applications \citep{DBLP:conf/ecir/BugertRBDG20}. 
To address these limitations, \citet{wec} proposed a large-scale, domain-independent cross-document coreference resolution dataset called WEC-Eng. 
Following this work, we carried out our study on cross-document event coreference resolution. 
However, the baseline model presented in WEC-Eng only considers naive event mention features for coreference resolution, overlooking rich discourse-level information in the document. Therefore, we propose modeling discourse-level information to enhance cross-document event coreference resolution.

Discourse information provides a comprehensive view of events.
On one hand, discourse structure can reveal the relationships between events and entities. 
The discourse structure often reflects certain logic clues such as causality and background,
which help in understanding the overall narrative flow and the relative positions of events.
As illustrated in Figure~\ref{fig1.b}, the phrases like ``four years later'' and ``before the head of  Hong Kong Special Administrative Region Election" correspond to EDU\ding{175} and EDU\ding{176}.
EDU\ding{175} provides background information for EDU\ding{176}.
Moreover, for long documents, there are many long-distance dependencies between sentences.
Due to the limitations of model input length, it is apt to cause information loss. 
Discourse structure information can shorten the distances between long-dependency sentences, providing a comprehensive understanding for the document.

Although discourse structure information is powerful, 
it cannot establish connections across documents. 
Therefore, semantic information beyond the discourse is required to establish fine-grained connections. 
Lexical information isolated in different documents 
sometimes cannot cover all the relevant information about the described events. 
Furthermore, the description of events in different documents may be quite different, resulting in the difficulty of event coreference resolution.
For instance, in Figure~\ref{fig1.a}, 
the same event might be represented differently such as ``Hong Kong" and ``Hong Kong Special Administrative Region", and the same meaning may be expressed using distinct words such as ``first" and ``inaugural".
Consequently, there is a need to establish connections between lexical information across cross-document contexts, allowing for the direct mapping of event descriptions and event arguments. 

In this work, we propose the DIE-EC (Discourse-Information-Enhanced Event Coreference) model, which leverages Rhetorical Structure Theory (RST) \citep{RST} to construct document-level RST trees and cross-document lexical chains \citep{lexical-chain}, aiming to model both the structural and semantic information within and cross documents. 
Our model consists of (1) an encoder layer, which encodes the input document to obtain contextual representations;
(2) a discourse information layer, which constructs document-level RST trees and cross-document-level lexical chains. The RST graph and lexical chains graph are processed using Graph Attention Networks (GAT) \citep{GAT};
(3) a pair scorer, which processes the results from the GAT using an MLP (Multi-Layer Perceptron) for coreference determination, and finally clusters events using the agglomerative clustering algorithm.
Besides the discourse information enhanced model, we also curate a Chinese cross-document event coreference dataset (WEC-Zh) due to its absence.
The WEC-Zh consists of 53,066 event mentions, 4,476 event clusters and 8 event types. 
It is the first large-scale  Chinese cross-document event coreference resolution dataset, promoting the research in the domains of Chinese and cross-lingual event coreference resolution.
We experimented our model on both WEC-Eng and WEC-Zh datasets, and it outperformed all baselines by large margins in various evaluation metrics.
To summarize, the contributions of this paper include:
\begin{enumerate}
\setlength{\itemsep}{0pt}
\setlength{\parsep}{0pt}
\setlength{\parskip}{0pt}
    \item We propose the DIE-EC model, which involves the joint graph of RST and Lexical Chains, and utilizes GAT to extract discourse and cross-document features from it.
    \item We introduce a large-scale Chinese cross-document event coreference resolution dataset to facilitate related research.
    \item We conduct experiments on both WEC-Eng and our proposed WEC-Zh datasets. Our model achieves the state-of-the-art results on both datasets.\footnote{Our code and dataset are available at \href{https://github.com/cooper12121/DIE-EC}{https://github.com/DIE-EC}.}
\end{enumerate}

\section{Related Work}
Prior studies on cross-document event coreference resolution are performed around several datasets such as ECB+ \citep{ecb+}, Gun Violence Corpus (GVC) \citep{GVC} and WEC-Eng \citep{wec}. We will introduce these datasets as well as their related methods as below.

\subsection{CDECR Datasets}
The ECB+ dataset is a cross-document event coreference resolution dataset with a relatively small scale, comprising only 982 samples. It encompasses 26,712 coreference links between 6,833 event mentions and 69,050 coreference links between 8,289 entity mentions.
Events are categorized by topic, comprising 43 topics in total. Each topic consists of different events that describe the same subject. Coreference annotations are specific to event categories and are only annotated within the same event category.

The GVC dataset pertains to the domain of gun violence.
 To overcome the costly endeavor of constructing large-scale datasets, the authors introduced an innovative semi-automatic method, known as structured-data-to-text (D2T). 
The D2T approach enables the creation of a vast amount of reference event data in a more efficient and highly consistent manner. 
GVC encompasses 5 gun-violence event classes (firing a gun, missing, hitting, injuring, death) and includes a total of 510 documents with 7,298 event mentions. 


WEC-Eng is a large-scale cross-document coreference resolution dataset. 
To address the limitations of existing datasets, i.e., dataset scales are small or coreference relations are determined only within the same event category, the authors proposed a new way of constructing datasets. 
Instead of requiring extensive annotations, only the validation and test sets underwent manual verification. 
WEC-Eng comprises 43,672 documents with a total of 7,597 clusters. 
The coreference determination in this dataset is independent of the event category, making it more aligned with real-world applications.
Although WEC-Eng has addressed the issues of existing datasets,
it primarily focuses on English and lacks data in other languages. Hence, we have constructed a large-scale Chinese dataset.

\vspace{-1mm}
\subsection{CDECR Methods}
Traditional coreference resolution methods mostly rely on mention representations for coreference score calculation. 
This approach often leads to a higher error rate when handling similar mentions. Therefore, 
\citet{corefqa} proposed determining coreference in a question-answering format. it only determines coreference for event entities, leading to less-than-optimal accuracy.
\citet{focus-on} introduced a coreference resolution model based on pre-filtering. This method reduces the computational cost by selecting only the top(n) mentions for coreference score calculation. However, the pre-filtering stage does introduce additional overhead.
 Similarly, \citet{task-dataset-modeling} built an end-to-end model based on the Deep Passage Retrieval (DPR) model, which consists of a retriever and a reader model.
 Given a query paragraph, the retriever selects the top k most relevant candidate paragraphs from the entire paragraph corpus.
\citet{pairwise} proposed a method that enhances the semantic representation of mentions by extracting event argument information to improve the accuracy of coreference judgments.  
\citet{HGCN} employed graph structures and multi-level iterative refinement to address coreference issues. 
The incorporation of syntactic parsing allows for fine-grained entity comparisons. 
However, the complexity of graph matrix construction is relatively high, making it unsuitable for long documents. 

Existing methods primarily rely on calculating mention similarities to determine coreference without harnessing comprehensive document-level information. As a result, they tend to underperform in the scenarios where mentions exhibit certain similarities, or mentions have different descriptions but similar meanings.
Therefore, we propose modeling document and cross-document information to enhance coreference resolution.

\section{DIE-EC: A Discourse-Information
-Enhanced Event Coreference Model for Cross-Document}

\begin{figure*}[!ht]
\begin{center}
\includegraphics[width=\textwidth,scale=1]{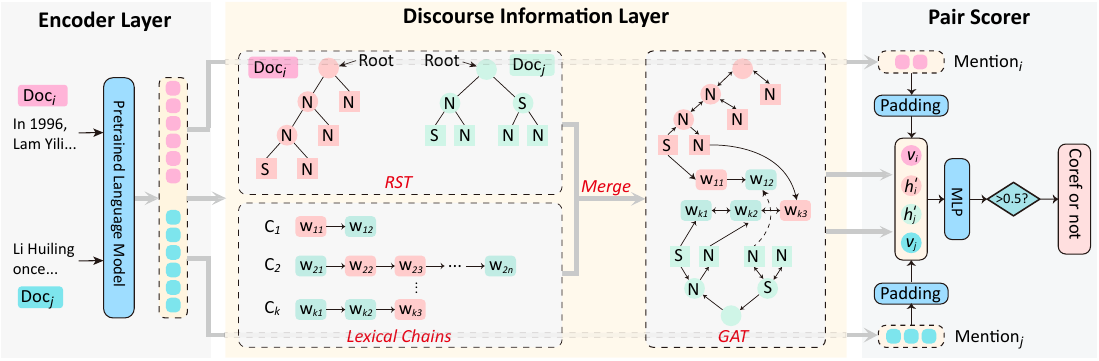} 
\caption{The architecture of DIE-EC. In the document-level RST trees, N denotes nucleus and S represents satellite. 
Lexical chains are across documents. 
After merging RST trees and lexical chains, Graph Attention Networks (GAT) is applied for representaton learning. 
Then an MLP is utilized to compute the coreference score.
}
\label{fig.3}
\end{center}
\end{figure*}

DIE-EC leverages document-level RST trees and cross-document lexical chains to model the structural and semantic information of documents, as shown in Figure ~\ref{fig.3}. 
For an input pair of documents, we first use Roberta \citep{roberta} to extract the semantic representation of event mentions. 
Simultaneously, we build document-level RST trees for each of the documents and lexical chains for multiple documents.
Then the RST trees are merged with lexical chains to obtain a comprehensive graph for the document pair. 
This combined graph is then fed into GAT for processing.
Then we employ the node representations from GAT and mention representations
to determine whether events are coreferential.
Finally, the co-referenced events are clustered using the agglomerative clustering algorithm. 

\subsection{Encoder Layer} 
For input documents \(doc_i\) and \(doc_j\), we separately employ Roberta to encode their contextual information, obtaining the contextual representations \(\bm{V}_i = \{\bm{v}_1,\cdots,\bm{v}_{n_i}\}\) and \(\bm{V}_j = \{\bm{v}_1,\cdots,\bm{v}_{n_j}\}\) for the documents. 
For event mentions \(i\) and \(j\) contained in the two documents, 
we first pad them to the same length and then use the concatenation of their token representations to represent them, denoted as \(\bm{v}_i\) and \(\bm{v}_j\) respectively.

\subsection{Discourse Information Layer}
\subsubsection{Cross-document Graph Construction}
\paragraph{RST Part}
We employ RST to represent the structural information of a document. 
Based on rhetorical relation theory, we analyze the discourse structure and function of the document, 
partitioning it into EDUs to build a rhetorical structure tree. 
The nodes in the tree have two types: nucleus and satellite.
The nucleus is a component containing the main information, while the satellite provides further explanation, supplementation, or modification to the nucleus. 
Different EDUs are connected by rhetorical relations. RST plays a crucial role in discourse comprehension. 
By analyzing the text's structure and rhetorical techniques, aids the model in better understanding the article's intent, the organization of information, and the relationship between various parts. 
We adopt the method proposed by \citet{zhang} to construct RST for the input documents, for Chinese data, we use the method proposed by \citet{CDTB} based on CDTB (Chinese Discourse Treebank) for construction. Let the nodes where event mentions $i$ and $j$ are located be represented as \( n_i \) and \( n_j \), respectively. In the RST tree, node relationships go from the satellite pointing to the nucleus or from the nucleus pointing to another nucleus. 
When converting the RST tree into a graph, we determine the direction of the edges in the graph based on the relationships between the nodes in the tree. 
Since the positions of two nucleus nodes are equivalent, we set the edge between the two nucleus nodes to be bidirectional.

\paragraph{Lexical Chain Part}
We utilize lexical chains to furnish semantic information of the document. Lexical chains offer insights that can assist systems in grasping the semantic connections between lexemes, thereby extracting information more precisely. 
It permits the system to account for semantic similarity, ensuring a more accurate determination of the relationships between entities or events mentioned within the document. 
Following the approach proposed by \citet{lexical-chains}, 
we construct lexical chains for repetition, synonyms, words close semantic relationships and temporal relationships. For the two input documents, we build cross-document lexical chains \( C_k = \{w_1,
w_2,\cdots,w_n\} {(k=1, \cdots, N)} \), where \( C_k \) denotes the $k$-th lexical chain constructed and \( w_n \) symbolizes the word within the chain. 
Let the lexical chains that contain the words from the event mentions of the two documents be \( C_i \) and \( C_j \), with starting lexemes \( w_1^i \) and \( w_1^j \) respectively. 
We construct a lexical chains graph with each word serving as a node, and all edges in the graph being bidirectional.

\paragraph{Merging RST and Lexical Chain} As only the leaf nodes in the RST graph contain the semantic information of the input document, we connect these leaf nodes with the nodes in the lexical chains graph to effectuate their fusion. 
The starting node of the lexical chain graph is connected to the EDU containing the starting lexeme; that is, \( w_i \) is linked to \( n_i \) and \( w_j \) is linked to \( n_j \). If different EDUs contain lexical chains, an edge is established between the corresponding EDUs. 
After merging, we obtain a connected graph, simultaneously endowed with both the document's structural and semantic information.

\subsubsection{Graph Representation Learning}
For the merged graph \( G \), we employ the GAT model to learn about the semantics and rhetorical relationships in the text, aiming to extract structural and semantic information. 
The GAT updates the feature representation of a target node by aggregating the features of its neighboring nodes. 
This approach eliminates the need to access information from the entire graph, enhancing the model's generalization capabilities. 
Moreover, through the attention mechanism, GAT can learn different attention weights for each neighboring node. 
This facilitates an effective aggregation of the neighboring nodes' features, leading to enhanced feature extraction capability for the target node.

For the graph \( G \) consisting of \( m \) nodes \(N=\{n_1, n_2,…, n_m\} \)
We use the representations derived from the encoder to initialize the nodes as 
\( \bm{h}=\{\bm{h}_1, \bm{h}_2, \cdots, \bm{h}_m\} \).
After processing through GAT, we obtain representations for the EDU nodes as
\( \bm{h}^{\prime}=\{\bm{h}^{\prime}_1, \bm{h}^{\prime}_2, \cdots, \bm{h}^{\prime}_m \}\), 
where $\bm{h}^{\prime}_i \in \mathbb{R}^{m\times d}$ and $d$ represents the dimension of node representations. \(\bm{h}^{\prime}_i \) can be calculated using the following formula:
\begin{equation}
\small
    \bm{h^{\prime}}_i =\sigma \left( \frac{1}{K} \sum_{i=1}^{K}\sum_{j \in \mathcal{N}(i)}  \frac{\exp(e_{ij})}{\sum_{k \in \mathcal{N}(i)}\exp(e_{ik})} \bm{W}_i \bm{h}_j \right)\,,
\end{equation}
where $K$ is the number of attention heads, $\sigma$ represents the Relu activation function, $\mathcal{N}(i)$ represents the set of neighboring nodes of node $i$, $\bm{W}_i$ is weight matrix for attention head.
$e_{ij}$ represents the attention weight between node $i$ and node $j$, which can be calculated as follows:
\begin{equation}
e_{ij} = \text{LeakyReLU}\left(\bm{a}_i^\top \left[\bm{W}_i\bm{h}_i || \bm{W}_i\bm{h}_j\right]\right)\,,
\end{equation}
where $\bm{a}_i$ is the attention parameter vector, $||$ represents the concatenation operation of vectors.
We use \(\bm{h}^{\prime}_i \) and \( \bm{h}^{\prime}_j \) to represent the EDU nodes where event mentions are located.

\subsection{Pair Scorer}
\paragraph{Computing Coreference Score} After obtaining the representations, we proceed to calculate the coreference score. 
We fuse the event mention representations \( \bm{v}_i \) and \( \bm{v}_j \) with the GAT node representations \( \bm{h}^{\prime}_i \) and \( \bm{h}^{\prime}_j \) to derive the representation for the event mention pair, denoted as:
\begin{equation}
\bm{v}(i,j)=[\bm{v}_i,\bm{v}_j,\bm{h}^{\prime}_i,\bm{h}^{\prime}_j]\,.
\end{equation}
Once we obtain the representation of the event mention pair, we employ a 3-layer MLP wherein the last layer performs a binary classification. 
Through the sigmoid function, we derive the coreference score for the mention pair as:
\begin{equation}
p\left(i,j\right)=sigmod\left(MLP\left(\bm{v}\left(i,j\right)\right)\right)\,.
\end{equation}

\paragraph{Clustering} All pairs with a probability lower than 0.5 are eliminated.
Following \citet{revisiting}, we use an agglomerative clustering algorithm for clustering.
It treats each data point as an initial cluster and progressively merges them into larger clusters. 
During the merging process, we use the group average, which measures the similarity between clusters based on the average distance of all elements within a cluster. 
The clustering process concludes when the similarity between all clusters falls below a predefined threshold of 0.7.

\section{WEC-Zh: Wikipedia Event Coreference Dataset for Chinese}
In this paper, we also curate a cross-document event coreference resolution dataset, WEC-Zh, to compensate the shortage of Chinese resource in this domain.
We will showcase the construction process and dataset statistics of the WEC-Zh dataset. 

\subsection{Construction Process}
Our dataset originates from the Wikipedia Chinese data. Following the precedent set by WEC-Eng \citep{wec}, our data was collected by aggregating anchor texts of Wikipedia links that point to the same Wikipedia concept. 
In other words, we consider links in an article that point to another article about the same real-world subject as being co-referential. 
Upon obtaining the raw data, we executed a series of filtering operations:
(1) Removing the mentions where event links were placed on parameter information (such as time, location, person and other entities that cannot represent events).
(2) Deleting non-Chinese event mentions.

\begin{figure}[!ht]
\begin{center}
\includegraphics[width=0.5\textwidth,scale=0.5]{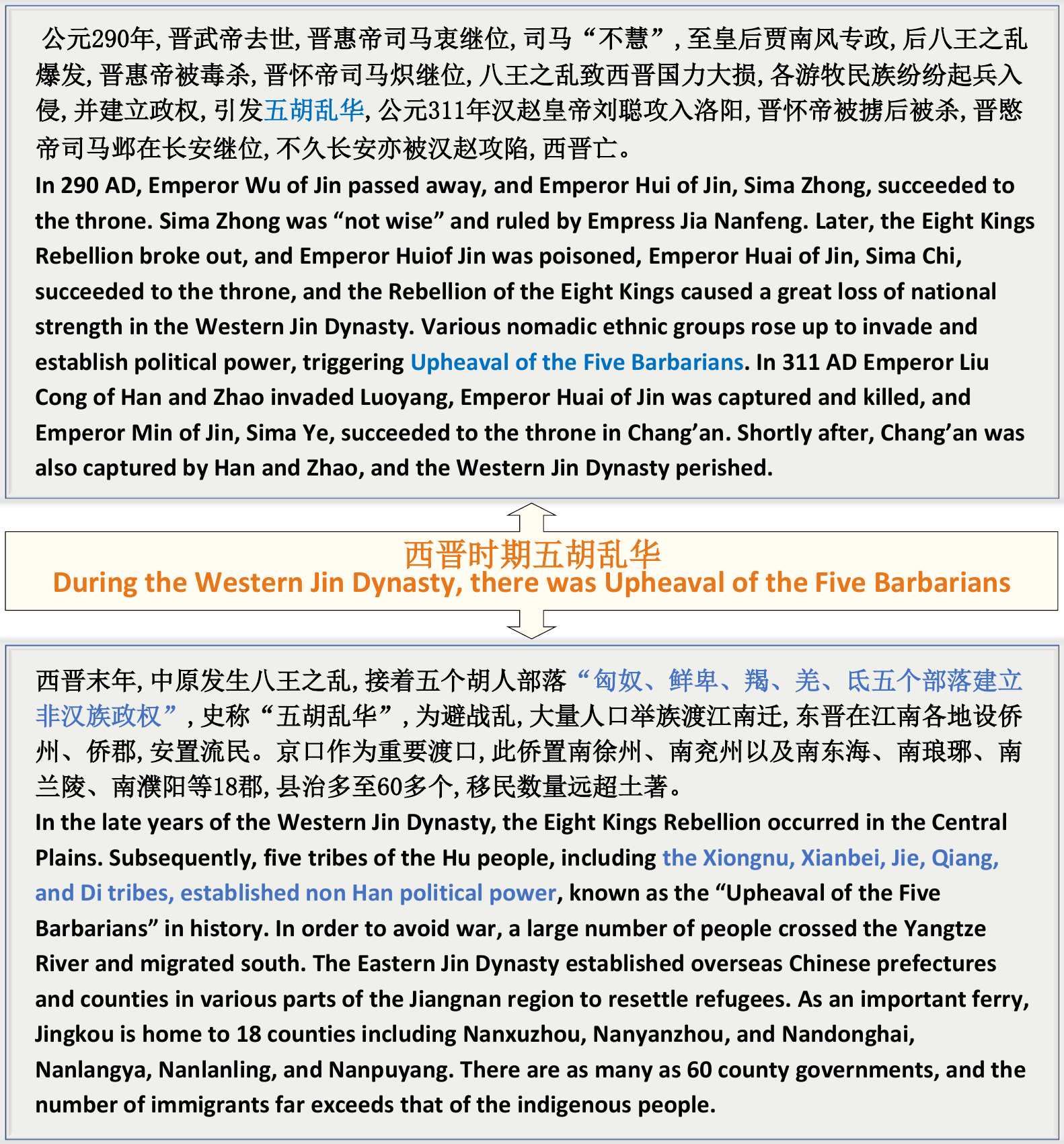} 
\caption{An example in WEC-Zh. \emph{\textcolor{myblue}{Upheaval of the Five Barbarians}} and \emph{\textcolor{myblue}{the Xiongnu, Xianbei, Jie, Qiang,
and Di tribes, established non-Han political power}} refer to the event ``During the Western Jin Dynasty, there was Upheaval of the Five Barbarians".
}
\label{chinese_sample}
\end{center}
\end{figure}

After the filtering operations, we converted Traditional Chinese to Simplified Chinese, ensuring that the resulting dataset consisting exclusively of Simplified Chinese data. 
To guarantee the quality of the dev and test sets, we conducted a two-person cross-validation on the dev and test datasets, eliminating noisy data and mentions whose coreferences could not be determined based on context. Figure \ref{chinese_sample} illustrates an example of WEC-zh.

To further assess the quality of annotators' labeling, we employed Fleiss' Kappa algorithm \citep{Kappa} to calculate annotation consistency. The Kappa consistency value is 0.71, indicating that our annotations achieves a acceptable agreement rate and showing the descent quality of our dataset.

\subsection{Dataset Statistics}
As shown in Table~\ref{table.1}, our dataset encompasses over 50,000 event mentions, which includes a certain number of singleton samples. 
Unlike traditional datasets like ECB+ \citep{ecb+} that are clustered by event categories and then coreference resolution is performed within each category, our dataset determines event coreference without such constraint.
To ensure domain diversity of the dataset, we tallied the numbers of documents with different event types, as presented in Table ~\ref{table.event_type}. 
For more detailed statistical information, please refer to Appendix~\ref{app:A}.

\begin{table}[ht]
\centering
\large
\resizebox{.5\textwidth}{!}{
\renewcommand{\arraystretch}{1.25}
\begin{tabular}{lccccc}
\hline
 & {Mentions} & {Clusters} & {Single Clusters} & {Ambiguity} & {Diversity} \\
\hline
{Train} & {49,861} & {3,855} & {1,110} & {3.31} & {2.35} \\
{Dev} & {1,538} & {297} & {84} & {1.82} & {1.69} \\
{Test} & {1,667} & {324} & {76} & {1.83} & {1.56} \\
\hline
\end{tabular}
}
\caption{\label{table.1}
WEC-Zh dataset statistics. `Single Clusters' are the clusters containing only one event mention, 
`Ambiguity' indicates the average number of different clusters in which a head lemma appears, and `Diversity' indicates the average number of unique head lemmas within a cluster.
}
\end{table}

\begin{table}[ht]
\centering
\small
\renewcommand{\arraystretch}{1.25}
\resizebox{.4\textwidth}{!}{
\begin{tabular}{lccc}
\hline
 & \textbf{Train} & \textbf{Dev} & \textbf{Test} \\
\hline
{ATTACK\_EVENT} & {31,569} & {886} & {917} \\
{SPORT\_EVENT} & {4,015} & {150} & {259}\\
{EVENT\_UNK} & {6,541} & {144} & {122}\\
{ELECTION\_EVENT} & {4,605} & {118} & {161}\\
{GENERAL\_EVENT} & {1,426} & {155} & {131}\\
{DISASTER\_EVENT} & {911} & {32} & {40}\\
{ACCIDENT\_EVENT} & {439} & {39} & {20}\\
{AWARD\_EVENT} & {197} & {12} & {15}\\
{OTHERS} & {158} & {2} & {2}\\
\hline
\end{tabular}
}
\caption{\label{table.event_type}
The numbers of different event types.}
\end{table}

\section{Experiments}

\begin{table*}[ht]
    \centering
    \small
    \resizebox{\textwidth}{!}{
    \begin{tabular}{@{}lccc ccc ccc c cccc@{}}\toprule
    & \phantom{abc}&\multicolumn{3}{c}{MUC} & \phantom{abc}& \multicolumn{3}{c}{$B^3$} & \phantom{abc}& \multicolumn{3}{c}{$CEAF$} & \phantom{abc}& CoNLL\\
    \cmidrule{3-5} \cmidrule{7-9} \cmidrule{11-13} \cmidrule{15-15}
    && R & P & $F_1$ && R & P & $F_1$ && R &P & $F_1$ && $F_1$  \\ 
    \midrule
        Lemma && \textbf{85.5} & 79.9 & \textbf{82.6} && 74.5 & 32.8 & 45.5 && 25.9 & 39.4 & 31.2 && 53.1 \\
        WEC-Eng && 78 & 83.6 & 80.7 && 66.1 & 55.3 & 60.2 && 53.4 & \textbf{40.3} & 45.9 && {62.3} \\
        Our model && 78.2 & \textbf{85.8} & 81.8 && 
        \textbf{69.6} & \textbf{62.4} & \textbf{65.8} && 
        \textbf{58.9} & 39.5 & \textbf{47.3} && 
        \textbf{65.0} \\
    \bottomrule
    \end{tabular}
    }
    \caption{Event coreference resolution results on the WEC-Eng test set.
    }
    \label{table.wec-eng}
\end{table*}

\begin{table*}[ht]
    \centering
    \small
    \resizebox{\textwidth}{!}{
    \begin{tabular}{@{}lcccccccccccccc@{}}\toprule
    & \phantom{abc}&\multicolumn{3}{c}{MUC} & \phantom{abc}& \multicolumn{3}{c}{$B^3$} & \phantom{abc}& \multicolumn{3}{c}{$CEAF$} & \phantom{abc}& CoNLL\\
    \cmidrule{3-5} \cmidrule{7-9} \cmidrule{11-13} \cmidrule{15-15}
    && R & P & $F_1$ && R & P & $F_1$ && R &P & $F_1$ && $F_1$  \\ 
    \midrule
        Lemma && 80.5 & 45.0 & 57.7 && 84.3 & 47.1 & 60.4 && 37.3 & 52.4 & 43.6 && 53.9 \\
        WEC-Eng && \textbf{91.1} & 54.2 & 68.0 && \textbf{91.9} & 52.3 & 66.7 && 39.0 & \textbf{76.3} & 51.6 && 62.1 \\
        Our model && 90.3 & \textbf{57.9} & \textbf{70.6}&& 90.8 & \textbf{60.2} & \textbf{72.4} && \textbf{46.7} & 76.0 & \textbf{57.8} && \textbf{66.9} \\
    \bottomrule
    \end{tabular}}
    \caption{Event coreference resolution results on the WEC-Zh test set}
    \label{table.wec-zh}
\end{table*}

\subsection{Experiment Setting}
\paratitle{Baselines}
We conducted experiments on both WEC-Eng and WEC-Zh.
The lemma model and WEC-Eng model proposed by \citet{wec} are used as baselines for both datasets.
In the lemma baseline, if two event mentions share the same syntactic-head lemma, they are considered co-referential.
To ensure a balanced sample, we maintained a positive-to-negative sample ratio of 1:10 for the training set. 
There was no such control for dev and test sets.
Positive samples consist of all mention pairs that belong to the same cluster. 

\paratitle{Evaluation Metrics}
We utilized the agglomerative clustering algorithm for clustering and reported P, R and $F_1$ scores on the MUC, $B_3$, CEAF, and CoNLL metrics for the clustering results.

\paratitle{Training Details}
Our experiments were conducted on two RTX 3090 GPUs. The initial learning rate was set to 1e-5 with a warm-up phase. The batch size was set to 128. We utilized the Adam optimizer as the optimization algorithm and set a weight decay parameter of 0.01 to prevent overfitting. 
The entire process was run for 10 iterations.

\subsection{Main Results}
\paratitle{WEC-Eng}
The experimental results of our model on WEC-Eng are shown in Table~\ref{table.wec-eng}. The results for the Lemma baseline and WEC-Eng baseline are sourced from the WEC-Eng. 
Our model outperforms the WEC-Eng baseline by 5.6 $F_1$ points of $B^3$, by 2.7 $F_1$ points of CoNLL, and also excels in the MUC metric compared to the WEC-Eng baseline. 
The WEC-Eng baseline has a higher error rate for mentions that are highly similar but not coreferential, whereas our model achieves a higher accuracy rate. 
This evidences that our model's adoption of RST and lexical chains for modeling the structural and semantic information of documents genuinely aids in better determining the coreference information of event mentions. 
The erroneous case studies are provided in Appendix~\ref{app:B}. 

\paratitle{WEC-Zh}
The experimental results on WEC-Zh are shown in Table~\ref{table.wec-zh}.
Our model outperforms the WEC-Eng model in terms of $F_1$ score across all metrics, with $F_1$ score of $B^3$ surpassing the WEC-Eng baseline by 5.7, and $F_1$ score of  CEAF exceeding the WEC-Eng baseline by 6.2. This suggests that, for the WEC-Zh dataset, our approach of constructing document-level RST trees and cross-document lexical chains better extracts event-related information from discourse-level text, to some extent reducing potential ambiguities and enhancing model performance.
Compared to the English dataset, our model achieved a more significant improvement on the Chinese dataset. 
This could be attributed to the fact that, in contrast to English, Chinese exhibits a wide variety of expressions for the same events, along with abbreviations and pronouns. 
This necessitates the model to establish correlations at the discourse-level for extracting useful information.

\begin{table*}[!]
    \centering
    {\fontsize{14}{12}\selectfont
    \renewcommand{\arraystretch}{2}
    \resizebox{
    \textwidth}{!}{
    \begin{tabular}{@{}lcccccccccccccc@{}}\toprule
    & \phantom{abc}&\multicolumn{3}{c}{MUC} & \phantom{abc}& \multicolumn{3}{c}{$B^3$} & \phantom{abc}& \multicolumn{3}{c}{$CEAF$} & \phantom{abc}& CoNLL\\
    \cmidrule{3-5} \cmidrule{7-9} \cmidrule{11-13} \cmidrule{15-15}
    && R & P & $F_1$ && R & P & $F_1$ && R &P & $F_1$ && $F_1$  \\ 
    \midrule
        Our-model 
             && 78.2 & {85.8} & 81.8 &&
              69.6 & {62.4} & {65.8} && 
                {58.9} & 39.5 & {47.3} && 
                {65.0} \\
        \texttt{-} RST 
             && 
             78.1 {\fontsize{10}{\baselineskip}\selectfont \textit{(-0.1)}} & 
             84.4 {\fontsize{10}{\baselineskip}\selectfont \textit{(-1.4)}} &              
             81.1 {\fontsize{10}{\baselineskip}\selectfont \textit{(-0.7)}}
             && 
             68.4 {\fontsize{10}{\baselineskip}\selectfont \textit{(-1.2)}}&
             60.1 {\fontsize{10}{\baselineskip}\selectfont \textit{(-2.3)}} &
             64.0 {\fontsize{10}{\baselineskip}\selectfont \textit{(-1.8)}}
             && 
             57.1 {\fontsize{10}{\baselineskip}\selectfont \textit{(-1.8)}}& 
             39.4 {\fontsize{10}{\baselineskip}\selectfont \textit{(-0.1)}} &
             46.6 {\fontsize{10}{\baselineskip}\selectfont \textit{(-0.7)}}
             && 63.9{\fontsize{10}{\baselineskip}\selectfont \textit{(-1.1)}}\\
        \texttt{-} Lexical chains 
            && 
            77.9{\fontsize{10}{\baselineskip}\selectfont \textit{(-0.3)}} &
            83.9 {\fontsize{10}{\baselineskip}\selectfont \textit{(-1.9)}} &
            80.8 {\fontsize{10}{\baselineskip}\selectfont \textit{(-1.0)}}
            && 
            64.2 {\fontsize{10}{\baselineskip}\selectfont \textit{(-5.2)}} &
            57.9 {\fontsize{10}{\baselineskip}\selectfont \textit{(-4.5)}} & 
            60.9 {\fontsize{10}{\baselineskip}\selectfont \textit{(-4.9)}}
            && 
            53.6 {\fontsize{10}{\baselineskip}\selectfont \textit{(-5.3)}} &
            39.2 {\fontsize{10}{\baselineskip}\selectfont \textit{(-0.3)}} &
            45.3 {\fontsize{10}{\baselineskip}\selectfont \textit{(-2.0)}}
            && 62.4 {\fontsize{10}{\baselineskip}\selectfont \textit{(-2.6)}} \\
    \bottomrule
    \end{tabular}
    }
    \caption{Ablation experiment results on the WEC-Eng test set.
    ``-RST" indicates the removal of the RST module, with only graph construction based on cross-document lexical chains, ``-Lexical chain" signifies the removal of lexical chains, with the focus solely on graph construction through RST processing.
    }
    \label{table.wec-eng_ablation}
    }
\end{table*}

\begin{table*}[!]
    \centering
    {\fontsize{14}{12}\selectfont
    \renewcommand{\arraystretch}{2.0}
    \resizebox{
    \textwidth}{!}{
    \begin{tabular}{@{}lcccccccccccccc@{}}\toprule
    & \phantom{abc}&\multicolumn{3}{c}{MUC} & \phantom{abc}& \multicolumn{3}{c}{$B^3$} & \phantom{abc}& \multicolumn{3}{c}{$CEAF$} & \phantom{abc}& CoNLL\\
    \cmidrule{3-5} \cmidrule{7-9} \cmidrule{11-13} \cmidrule{15-15}
    && R & P & $F_1$ && R & P & $F_1$ && R &P & $F_1$ && $F_1$  \\ 
    \midrule
        Our-model 
             && 90.3 & {57.9} &{70.6}&& 90.8 & {60.2} &{72.4} && {46.7} & 76.0 & {57.8} && {66.9} \\
        \texttt{-} RST 
             && 
             90.2 {\fontsize{10}{\baselineskip}\selectfont \textit{(-0.1)}} & 
             55.3 {\fontsize{10}{\baselineskip}\selectfont \textit{(-2.6)}} & 
             68.6 {\fontsize{10}{\baselineskip}\selectfont \textit{(-2.0)}}
             && 
             89.8 {\fontsize{10}{\baselineskip}\selectfont \textit{(-1.0)}}&
             55.8 {\fontsize{10}{\baselineskip}\selectfont \textit{(-4.4)}} &
             68.8    {\fontsize{10}{\baselineskip}\selectfont \textit{(-3.6)}}
             && 
             42.6 {\fontsize{10}{\baselineskip}\selectfont \textit{(-0.1)}}& 
             75.8 {\fontsize{10}{\baselineskip}\selectfont \textit{(-0.2)}} &
             54.5 {\fontsize{10}{\baselineskip}\selectfont \textit{(-3.3)}}
             &&  
             64.0{\fontsize{10}{\baselineskip}\selectfont \textit{(-2.9)}}\\
        \texttt{-} Lexical chains 
            && 
            90.1{\fontsize{10}{\baselineskip}\selectfont \textit{(-0.2)}} &
            54.7 {\fontsize{10}{\baselineskip}\selectfont \textit{(-3.2)}} &
            68.1 {\fontsize{10}{\baselineskip}\selectfont \textit{(-2.5)}}
            && 
            87.4 {\fontsize{10}{\baselineskip}\selectfont \textit{(-3.4)}} &
            54.1 {\fontsize{10}{\baselineskip}\selectfont \textit{(-6.1)}} & 
            66.8 {\fontsize{10}{\baselineskip}\selectfont \textit{(-5.6)}}
            && 
            40.5 {\fontsize{10}{\baselineskip}\selectfont \textit{(-6.2)}} &
            75.9 {\fontsize{10}{\baselineskip}\selectfont \textit{(-0.1)}} &
            53.0 {\fontsize{10}{\baselineskip}\selectfont \textit{(-5.0)}}
            && 62.6 {\fontsize{10}{\baselineskip}\selectfont \textit{(-4.3)}} \\
    \bottomrule
    \end{tabular}
    }
    \caption{Ablation experiment results on the WEC-Zh test set.
    }
    \label{table.wec-zh-ablation}
    }
\end{table*}
\subsection{Ablation Studies}
To further investigate the contribution of RST trees and lexical chains for cross-document coreference resolution, we conducted a series of ablation experiments, and the experimental results are shown in Table \ref{table.wec-eng_ablation} and \ref{table.wec-zh-ablation}.

From the ablation study results, it is evident that removing either RST trees or lexical chains leads to a decrease in model performance. 
This indicates that discourse information indeed enhances the model's coreference resolution capabilities. 
This is attributed to the fact that the discourse structure provides the logical relationships between an event and its surrounding context, while the discourse semantics captures the continuity of the event and the evolutionary process of related information. 
These two types of information complement each other, aiding the system in a deeper understanding and comparison of event descriptions across different documents.

Additionally, a significant performance drop is observed upon removing the lexical chains, suggesting that even though the RST trees provide structural information of the discourse, the contribution of lexical chains to the model's performance is significant. 
This might indicate that in cross-document event coreference resolution tasks, deep semantic coherence and topic continuity play a pivotal role. While the structural relationships provided by RST trees do assist in coreference resolution, they are not as direct and crucial as the semantic coherence captured by lexical chains. 
When dealing with events across documents, recognizing and matching the same or similar semantic content might be more important than just understanding the structure and narrative flow.

\subsection{In-depth Analysis}
\begin{figure}
\centering
\resizebox{0.5\textwidth}{!}{ 
\begin{subfigure}{0.5\textwidth}
 \centering
\begin{tikzpicture}
  \begin{axis}[
    xlabel={lexical overlap rate (\%)},
    ylabel={$F_1$},
    xticklabel style={font=\large},
    yticklabel style={font=\large},
    xtick={0,30,65,100}, 
    xticklabels={0-10,10-30,30-50,50-100},
    ytick={60,70,80}, 
    yticklabels={60,70,80},
    ymin=40, 
    ymax=90,
    xmin=0,
    xmax=100,
    ybar, 
    bar width=15pt, 
    nodes near coords, 
    nodes near coords align={vertical}, 
    every node near coord/.append style={font=\fontsize{11}{12}\selectfont}, 
    axis x line=bottom, 
    axis y line=left, 
    legend style={
    at={(0.5,1.0)}, anchor=north,legend columns=-1,
    row sep = 0.1em,
    column sep = 1em,
    },
    enlarge x limits=0.2,
    xlabel style={
        font=\large
    },
    ylabel style={
        xshift = 2.5cm,
        yshift = -0.8cm,
        font = \large
    },
  ]
  
  \addplot[fill=overlap_full] coordinates {(0, 58.8) (30, 64.3) (65, 67.4) (100, 71.3)};
  \addplot[fill=overlap_lexi] coordinates {(1, 50.3) (31, 60.1) (66, 65.2) (101, 70.5)};

  \legend{full-model, - lexical chains}
  \end{axis}
\end{tikzpicture}
\captionsetup{font=large}
\caption{Results on WEC-Eng}
\label{impact_of_lexical_overlap on WEC-Eng}
\end{subfigure}

\hfill
\begin{subfigure}{0.5\textwidth}
 \centering
   \begin{tikzpicture}
  \begin{axis}[
    xlabel={lexical overlap rate (\%)},
    ylabel={$F_1$},
    xticklabel style={font=\large},
    yticklabel style={font=\large},
    xtick={0,30,65,100}, 
    xticklabels={0-10,10-30,30-50,50-100},
     ytick={60,70,80}, 
    yticklabels={60,70,80},
    ymin=40, 
    ymax=90,
    xmin=0,
    xmax=100,
    ybar, 
    bar width=15pt, 
    nodes near coords, 
    nodes near coords align={vertical}, 
    every node near coord/.append style={font=\fontsize{11}{10}\selectfont}, 
    axis x line=bottom, 
    axis y line=left, 
    legend style={
    at={(0.5,1.0)}, anchor=north,legend columns=-1,
    row sep = 0.1em,
    column sep = 1em,
    },
    enlarge x limits=0.2,
    xlabel style={
        font=\large
    },
    ylabel style={
        xshift = 2.5cm,
        yshift = -0.8cm,
        font = \large
    },
  ]
  
  \addplot[fill=overlap_full] coordinates {(0, 67.0) (30, 71.1) (65, 76.5) (100, 80.1)};
  \addplot[fill=overlap_lexi] coordinates {(1, 56.9) (31, 64.6) (66, 73.3) (101, 79.2)};

  \legend{full-model, - lexical chains}
  \end{axis}
\end{tikzpicture}
\captionsetup{font=large}
\caption{Results on WEC-Zh}
\label{impact_of_lexical_overlap on WEC-Zh}
\end{subfigure}

}
\caption{The impact of lexical chains with different lexical overlap rates.
}\label{fig.impact of lexical overlap rate}
\end{figure}
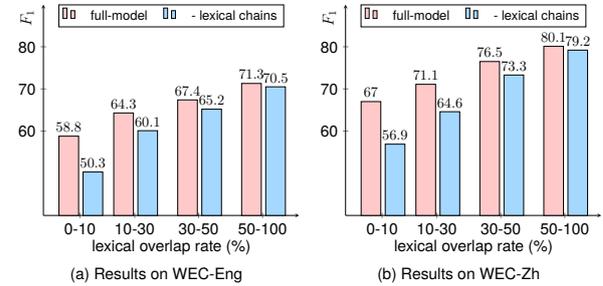

To further analyze the effects of our RST trees and lexical chains on different samples, we conducted two experiments: (1) The impact of lexical chains with different lexical overlap rates. The lexical overlap rate is calculated as the ratio of overlapping words between two documents to the average total number of words in both documents.
(2) The impact of RST tree with different document lengths. The length of a document is defined as the number of words it contains.
\paragraph{(1) Different Lexical Overlap Rates}
We hypothesized that events which are likely to be coreferent often contain a higher degree of lexical overlap between documents, as their descriptions share similar content. 
In such cases, the model can make judgments more easily, and the role of lexical chains is relatively minor. Conversely, for non-coreferent events, where descriptions differ or are only partially similar, the lexical overlap between documents is lower, leading to potential confusion for the model. 
In these cases, lexical chains play a more critical role in connecting vocabulary across documents to assist the model in making judgments. 
To validate our hypothesis, we computed the lexical overlap rates for all sample pairs and conducted ablation experiments, measuring $F_1$ scores of $B^3$. The experimental results are shown in Figure \ref{fig.impact of lexical overlap rate}. 
It is evident that when the lexical overlap rate exceeds 50\%, it is easier to determine co-reference, removing lexical chains does not lead to a significant decrease in the model's performance. However, when the overlap rate falls below 30\%, the model's performance starts to decline substantially. Specifically, when it drops below 10\%, $F_1$ scores decrease by 8.5 on WEC-Eng, and by 10.1 on WEC-Zh. This validates our hypothesis that for texts with a low lexical overlap rate, lexical chains play a significant role in co-reference resolution.

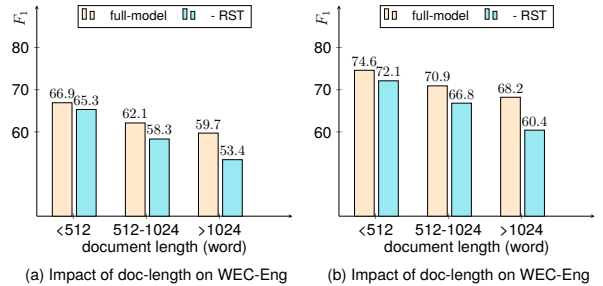
\begin{figure}
  \centering  
\resizebox{0.5\textwidth}{!}{ 

\begin{subfigure}{0.5\textwidth}
\begin{tikzpicture}
  \begin{axis}[
    xlabel={document length (word)},
    ylabel={$F_1$},
    xticklabel style={font=\large},
    yticklabel style={font=\large},
    xtick={0,40,80}, 
    xticklabels={<512,512-1024,>1024},
    ytick={60,70,80}, 
    yticklabels={60,70,80},
    ymin=40, 
    ymax=90,
    xmin=0,
    xmax=100,
    ybar, 
    bar width=15pt, 
    nodes near coords, 
    nodes near coords align={vertical}, 
    every node near coord/.append style={font=\fontsize{11}{12}\selectfont}, 
    axis x line=bottom, 
    axis y line=left, 
    legend style={
    at={(0.5,1.0)}, anchor=north,legend columns=-1,
    row sep = 0.1em,
    column sep = 1em,
    },
    enlarge x limits=0.2,
    xlabel style={
       font = \large
    },
    ylabel style={
        xshift = 2.5cm,
        yshift = -0.8cm,
        font = \large
    },
  ]
  
  \addplot[fill=doclength_full] coordinates {(0, 66.9) (40, 62.1) (80, 59.7) };
  \addplot[fill=doclength_rst] coordinates {(1, 65.3) (41, 58.3) (81, 53.4)};

  \legend{full-model, - RST}
  
  \end{axis}
\end{tikzpicture}
\captionsetup{font=large}
\caption{Impact of doc-length on WEC-Eng}
\label{impact_of_doclength on wec-eng}    
\end{subfigure}
\hfill
\begin{subfigure}{0.5\textwidth}
\begin{tikzpicture}
  \begin{axis}[
    xlabel={document length (word)},
    ylabel={$F_1$},
    xticklabel style={font=\large},
    yticklabel style={font=\large},
    xtick={0,40,80}, 
    xticklabels={<512,512-1024,>1024},
    ytick={60,70,80}, 
    yticklabels={60,70,80},
    ymin=40, 
    ymax=90,
    xmin=0,
    xmax=100,
    ybar, 
    bar width=15pt, 
    nodes near coords, 
    nodes near coords align={vertical}, 
    every node near coord/.append style={font=\fontsize{11}{12}\selectfont}, 
    axis x line=bottom, 
    axis y line=left, 
    legend style={
    at={(0.5,1.0)}, anchor=north,legend columns=-1,
    row sep = 0.1em,
    column sep = 1em,
    },
    enlarge x limits=0.2,
    xlabel style={
       font = \large
    },
    ylabel style={
        xshift = 2.5cm,
        yshift = -0.8cm,
        font = \large
    },
  ]
  
  \addplot[fill=doclength_full] coordinates {(0, 74.6) (40, 70.9) (80, 68.2) };
  \addplot[fill=doclength_rst] coordinates {(1, 72.1) (41, 66.8) (81, 60.4)};

  \legend{full-model, - RST}
  
  \end{axis}
\end{tikzpicture}
\captionsetup{font=large}
\caption{Impact of doc-length on WEC-Eng}
\label{impact_of_doclength on wec-eng}    
\end{subfigure}
}
\caption{The impact of RST with different document lengths.}\label{impact of document length}
\end{figure}

\begin{table*}[t]
    \centering
    \small
    \resizebox{\textwidth}{!}{
    \begin{tabular}{@{}lccc ccc ccc c cccc@{}}\toprule
    & \phantom{abc}&\multicolumn{3}{c}{MUC} & \phantom{abc}& \multicolumn{3}{c}{$B^3$} & \phantom{abc}& \multicolumn{3}{c}{$CEAF$} & \phantom{abc}& CoNLL\\
    \cmidrule{3-5} \cmidrule{7-9} \cmidrule{11-13} \cmidrule{15-15}
    && R & P & $F_1$ && R & P & $F_1$ && R &P & $F_1$ && $F_1$  \\ 
    \midrule
        PairwiseRL && 
        88.1 & 85.1 & 86.6 &&
        86.1 & 84.7 & 85.4 && 
        83.1 & 79.6 & 81.3 && 
        84.4 \\
        DRS CD-ECR &&
        \textbf{88.6} & 85.9 & 87.2 && 
        \textbf{87.8} & 85.4 & \textbf{86.6} && 
        \textbf{82.8} & \textbf{83.7} & \textbf{83.2} && 
        \textbf{85.7} \\
        Our model && 88.1 & \textbf{87.6} & \textbf{87.8} && 
        86.9 & \textbf{85.8} & 86.3 && 
        81.7 & 83.2 & 82.4&& 
        85.5 \\
    \bottomrule
    \end{tabular}
    }
    \caption{Event coreference resolution results on the ECB+ test set.
    }
    \label{table.ecb+}
\end{table*}

\begin{table*}[ht]
    \centering
    {\fontsize{14}{12}\selectfont
    \renewcommand{\arraystretch}{2}
    \resizebox{
    \textwidth}{!}{
    \begin{tabular}{@{}lcccccccccccccc@{}}\toprule
    & \phantom{abc}&\multicolumn{3}{c}{MUC} & \phantom{abc}& \multicolumn{3}{c}{$B^3$} & \phantom{abc}& \multicolumn{3}{c}{$CEAF$} & \phantom{abc}& CoNLL\\
    \cmidrule{3-5} \cmidrule{7-9} \cmidrule{11-13} \cmidrule{15-15}
    && R & P & $F_1$ && R & P & $F_1$ && R &P & $F_1$ && $F_1$  \\ 
    \midrule
        Our model 
        &&  88.1 & 87.6 & 87.8 && 
        86.9 & 85.8 & 86.3 && 
        81.7 & 83.2 & 82.4&& 
        85.5 \\
        \texttt{-} RST 
             && 
             86.3 {\fontsize{10}{\baselineskip}\selectfont \textit{(-1.8)}} & 
             84.2 {\fontsize{10}{\baselineskip}\selectfont \textit{(-3.4)}} & 
             85.2 {\fontsize{10}{\baselineskip}\selectfont \textit{(-2.6)}}
             && 
             85.1 {\fontsize{10}{\baselineskip}\selectfont \textit{(-1.8)}}&
             83.5 {\fontsize{10}{\baselineskip}\selectfont \textit{(-2.3)}} &
             84.3    {\fontsize{10}{\baselineskip}\selectfont \textit{(-2.0)}}
             && 
             79.8 {\fontsize{10}{\baselineskip}\selectfont \textit{(-1.9)}}& 
             82.1 {\fontsize{10}{\baselineskip}\selectfont \textit{(-1.1)}} &
             81.0 {\fontsize{10}{\baselineskip}\selectfont \textit{(-1.4)}}
             &&  
             83.5{\fontsize{10}{\baselineskip}\selectfont \textit{(-2.0)}}\\
        \texttt{-} Lexical chains 
            && 
            84.9{\fontsize{10}{\baselineskip}\selectfont \textit{(-3.2)}} &
            83.7 {\fontsize{10}{\baselineskip}\selectfont \textit{(-3.9)}} &
            84.3 {\fontsize{10}{\baselineskip}\selectfont \textit{(-3.5)}}
            && 
            84.2 {\fontsize{10}{\baselineskip}\selectfont \textit{(-2.7)}} &
            82 {\fontsize{10}{\baselineskip}\selectfont \textit{(-3.8)}} & 
            83.1 {\fontsize{10}{\baselineskip}\selectfont \textit{(-3.2)}}
            && 
            78.3 {\fontsize{10}{\baselineskip}\selectfont \textit{(-3.4)}} &
            80.4 {\fontsize{10}{\baselineskip}\selectfont \textit{(-2.8)}} &
            79.3 {\fontsize{10}{\baselineskip}\selectfont \textit{(-3.1)}}
            && 82.2 {\fontsize{10}{\baselineskip}\selectfont \textit{(-3.3)}} \\
    \bottomrule
    \end{tabular}
    }
    \caption{Ablation experiment results on the ECB+ test set.
    }
    \label{table.ecb+-ablation}
    }
\end{table*}

\paragraph{(2) Different Document Lengths}
Longer documents often contain a wealth of contextual dependencies. Previous methods, due to direct context truncation, may overlook this information, resulting in some loss of information. We hypothesize that RST trees can model the discourse structure of long documents, extracting long-range dependencies that enhance coreference resolution. To validate our hypothesis, we conducted experiments on document pairs of varying lengths. The experimental results are shown in Figure \ref{impact of document length}. It is evident that when the document length is less than 512, removing the RST model only results in a slight performance drop. However, when the document length exceeds 1024, the F1 scores decrease by 6.3  on WEC-Eng, by 7.8 on WEC-Zh. This demonstrates that RST is capable of modeling long documents and capturing long-range dependencies, thus validating our hypothesis.


\subsection{ECB+ Experiment}\label{ECB+ Experiment}

Besides the experiments on WEC-Eng and WEC-Zh, we also conducted experiments on ECB+ \citep{ecb+} because it is widely used for evaluating models of cross-document event coreference resolution.

\subsubsection{Experimental Settings}
When forming mention pairs from the ECB+ dataset, all positive samples are paired with each other, and all possible negative samples within the same subtopic are also paired with each other. The final distribution of these pairs is as follows:

\textit{Dev set: Positives=5,881, Negatives=50,653}

\textit{Test set: Positives=6,889, Negatives=87,053}

\textit{Train set: Positives=14,944, Negatives=170,549}

\noindent To show the effectiveness of our model, we selected two SOTA models proposed for ECB+ as baselines, namely PairwiseRL \cite{pairwise} and DRS CD-ECR \cite{chen-etal-2023-cross-discourse}.

\subsubsection{Experimental Results}
The results of the baseline comparison experiment and ablation studies are shown in Table \ref{table.ecb+} and \ref{table.ecb+-ablation} respectively.
From the experimental results, it can be seen that our model achieves comparable performance with the SOTA model DRS CD-ECR, which is reasonable because both of our and their models utilize discourse structural knowledge. 
The performance differences may be due to the implementation details such as using different RST parsers or the methods of leveraging RST.
Compared to PairwiseRL, our model surpasses it in MUC by 1.2, B3 by 0.9, CEAF by 1.1 and CoNLL by 1.1.
This demonstrates that our approach of modeling the global structural and semantic information of documents is more effective than enhancing event mention representations by extracting event parameters.

\section{Conclusion}
In this study, we introduce an improved method for cross-document event coreference resolution by incorporating discourse-level information.
By constructing document-level RST trees and cross-document Lexical chains, we can more effectively capture the structural and semantic information within documents, especially long-distance dependencies. Moreover, we have constructed a large-scale cross-document event coreference dataset for the Chinese context, filling a gap in current research. 
This work offers valuable resources and methodologies for further studies on cross-document event coreference resolution.


\section{Bibliographical References}
\bibliographystyle{lrec-coling2024-natbib}
\bibliography{lrec-coling2024-example}

\begin{thebibliography}{24}
\expandafter\ifx\csname natexlab\endcsname\relax\def\natexlab#1{#1}\fi

\bibitem[{Barhom et~al.(2019)Barhom, Shwartz, Eirew, Bugert, Reimers, and Dagan}]{revisiting}
Shany Barhom, Vered Shwartz, Alon Eirew, Michael Bugert, Nils Reimers, and Ido Dagan. 2019.
\newblock \href {https://doi.org/10.18653/v1/P19-1409} {Revisiting joint modeling of cross-document entity and event coreference resolution}.
\newblock In \emph{Proceedings of the 57th Annual Meeting of the Association for Computational Linguistics}, pages 4179--4189, Florence, Italy. Association for Computational Linguistics.

\bibitem[{Bugert et~al.(2020)Bugert, Reimers, Barhom, Dagan, and Gurevych}]{DBLP:conf/ecir/BugertRBDG20}
Michael Bugert, Nils Reimers, Shany Barhom, Ido Dagan, and Iryna Gurevych. 2020.
\newblock \href {https://ceur-ws.org/Vol-2593/paper3.pdf} {Breaking the subtopic barrier in cross-document event coreference resolution}.
\newblock In \emph{Proceedings of Text2Story - Third Workshop on Narrative Extraction From Texts co-located with 42nd European Conference on Information Retrieval, Text2Story@ECIR 2020, Lisbon, Portugal, April 14th, 2020 [online only]}, volume 2593 of \emph{{CEUR} Workshop Proceedings}, pages 23--29. CEUR-WS.org.

\bibitem[{Chen et~al.(2023)Chen, Xu, Li, and Zhu}]{chen-etal-2023-cross-discourse}
Xinyu Chen, Sheng Xu, Peifeng Li, and Qiaoming Zhu. 2023.
\newblock \href {https://doi.org/10.18653/v1/2023.emnlp-main.294} {Cross-document event coreference resolution on discourse structure}.
\newblock In \emph{Proceedings of the 2023 Conference on Empirical Methods in Natural Language Processing}, pages 4833--4843, Singapore. Association for Computational Linguistics.

\bibitem[{Cybulska and Vossen(2014)}]{ecb+}
Agata Cybulska and Piek Vossen. 2014.
\newblock \href {http://www.lrec-conf.org/proceedings/lrec2014/pdf/840_Paper.pdf} {Using a sledgehammer to crack a nut? lexical diversity and event coreference resolution}.
\newblock In \emph{Proceedings of the Ninth International Conference on Language Resources and Evaluation ({LREC}'14)}, pages 4545--4552, Reykjavik, Iceland. European Language Resources Association (ELRA).

\bibitem[{Dobrovolskii(2021)}]{word-level}
Vladimir Dobrovolskii. 2021.
\newblock \href {https://doi.org/10.18653/v1/2021.emnlp-main.605} {Word-level coreference resolution}.
\newblock In \emph{Proceedings of the 2021 Conference on Empirical Methods in Natural Language Processing}, pages 7670--7675, Online and Punta Cana, Dominican Republic. Association for Computational Linguistics.

\bibitem[{Eirew et~al.(2022)Eirew, Caciularu, and Dagan}]{task-dataset-modeling}
Alon Eirew, Avi Caciularu, and Ido Dagan. 2022.
\newblock \href {https://doi.org/10.18653/v1/2022.emnlp-main.58} {Cross-document event coreference search: Task, dataset and modeling}.
\newblock In \emph{Proceedings of the 2022 Conference on Empirical Methods in Natural Language Processing}, pages 900--913, Abu Dhabi, United Arab Emirates. Association for Computational Linguistics.

\bibitem[{Eirew et~al.(2021)Eirew, Cattan, and Dagan}]{wec}
Alon Eirew, Arie Cattan, and Ido Dagan. 2021.
\newblock \href {https://doi.org/10.18653/v1/2021.naacl-main.198} {{WEC}: Deriving a large-scale cross-document event coreference dataset from {W}ikipedia}.
\newblock In \emph{Proceedings of the 2021 Conference of the North American Chapter of the Association for Computational Linguistics: Human Language Technologies}, pages 2498--2510, Online. Association for Computational Linguistics.

\bibitem[{Fleiss(1971)}]{Kappa}
JL~Fleiss. 1971.
\newblock \href {https://doi.org/10.1037/h0031619} {Measuring nominal scale agreement among many raters}.
\newblock \emph{Psychological bulletin}, 76(5):378—382.

\bibitem[{Held et~al.(2021)Held, Iter, and Jurafsky}]{focus-on}
William Held, Dan Iter, and Dan Jurafsky. 2021.
\newblock \href {https://doi.org/10.18653/v1/2021.emnlp-main.106} {Focus on what matters: Applying discourse coherence theory to cross document coreference}.
\newblock In \emph{Proceedings of the 2021 Conference on Empirical Methods in Natural Language Processing}, pages 1406--1417, Online and Punta Cana, Dominican Republic. Association for Computational Linguistics.

\bibitem[{Khattab et~al.(2021)Khattab, Potts, and Zaharia}]{baleen}
Omar Khattab, Christopher Potts, and Matei Zaharia. 2021.
\newblock Baleen: Robust multi-hop reasoning at scale via condensed retrieval.
\newblock \emph{Advances in Neural Information Processing Systems}, 34:27670--27682.

\bibitem[{Kong and Zhou(2017)}]{CDTB}
Fang Kong and Guodong Zhou. 2017.
\newblock \href {https://doi.org/10.1145/3099557} {A cdt-styled end-to-end chinese discourse parser}.
\newblock \emph{ACM Trans. Asian Low-Resour. Lang. Inf. Process.}, 16(4).

\bibitem[{Li et~al.(2020)Li, Xiao, Liu, Wu, Wang, and Du}]{leveraging}
Wei Li, Xinyan Xiao, Jiachen Liu, Hua Wu, Haifeng Wang, and Junping Du. 2020.
\newblock Leveraging graph to improve abstractive multi-document summarization.
\newblock \emph{arXiv preprint arXiv:2005.10043}.

\bibitem[{Liu et~al.(2019)Liu, Ott, Goyal, Du, Joshi, Chen, Levy, Lewis, Zettlemoyer, and Stoyanov}]{roberta}
Yinhan Liu, Myle Ott, Naman Goyal, Jingfei Du, Mandar Joshi, Danqi Chen, Omer Levy, Mike Lewis, Luke Zettlemoyer, and Veselin Stoyanov. 2019.
\newblock \href {http://arxiv.org/abs/1907.11692} {Roberta: A robustly optimized bert pretraining approach}.

\bibitem[{Mann and Thompson(1987)}]{RST}
William Mann and Sandra Thompson. 1987.
\newblock \emph{Rhetorical Structure Theory: A Theory of Text Organization}.

\bibitem[{Miculicich and Henderson(2022)}]{HGCN}
Lesly Miculicich and James Henderson. 2022.
\newblock \href {https://doi.org/10.18653/v1/2022.findings-acl.215} {Graph refinement for coreference resolution}.
\newblock In \emph{Findings of the Association for Computational Linguistics: ACL 2022}, pages 2732--2742, Dublin, Ireland. Association for Computational Linguistics.

\bibitem[{Morris and Hirst(1991)}]{lexical-chain}
Jane Morris and Graeme Hirst. 1991.
\newblock \href {https://aclanthology.org/J91-1002} {Lexical cohesion computed by thesaural relations as an indicator of the structure of text}.
\newblock \emph{Computational Linguistics}, 17(1):21--48.

\bibitem[{Ravenscroft et~al.(2021)Ravenscroft, Clare, Cattan, Dagan, and Liakata}]{CD2CR}
James Ravenscroft, Amanda Clare, Arie Cattan, Ido Dagan, and Maria Liakata. 2021.
\newblock \href {https://doi.org/10.18653/v1/2021.eacl-main.21} {{CD}{\^{}}2{CR}: Co-reference resolution across documents and domains}.
\newblock In \emph{Proceedings of the 16th Conference of the European Chapter of the Association for Computational Linguistics: Main Volume}, pages 270--280, Online. Association for Computational Linguistics.

\bibitem[{Ruas et~al.(2020)Ruas, Ferreira, Grosky, de~França, and de~Medeiros}]{lexical-chains}
Terry Ruas, Charles Henrique~Porto Ferreira, William~I. Grosky, Fabr{\'i}cio~Olivetti de~França, and Debora Maria~Rossi de~Medeiros. 2020.
\newblock \href {https://api.semanticscholar.org/CorpusID:218954068} {Enhanced word embeddings using multi-semantic representation through lexical chains}.
\newblock \emph{ArXiv}, abs/2101.09023.

\bibitem[{Veličković et~al.(2018)Veličković, Cucurull, Casanova, Romero, Liò, and Bengio}]{GAT}
Petar Veličković, Guillem Cucurull, Arantxa Casanova, Adriana Romero, Pietro Liò, and Yoshua Bengio. 2018.
\newblock \href {http://arxiv.org/abs/1710.10903} {Graph attention networks}.

\bibitem[{Vossen et~al.(2018)Vossen, Ilievski, Postma, and Segers}]{GVC}
Piek Vossen, Filip Ilievski, Marten Postma, and Roxane Segers. 2018.
\newblock \href {https://aclanthology.org/L18-1480} {Don{'}t annotate, but validate: a data-to-text method for capturing event data}.
\newblock In \emph{Proceedings of the Eleventh International Conference on Language Resources and Evaluation ({LREC} 2018)}, Miyazaki, Japan. European Language Resources Association (ELRA).

\bibitem[{Wen et~al.(2021)Wen, Lin, Lai, Pan, Li, Lin, Zhou, Li, Wang, Zhang, Yu, Dong, Wang, Fung, Mishra, Lyu, Sur{\'\i}s, Chen, Brown, Palmer, Callison-Burch, Vondrick, Han, Roth, Chang, and Ji}]{resin}
Haoyang Wen, Ying Lin, Tuan Lai, Xiaoman Pan, Sha Li, Xudong Lin, Ben Zhou, Manling Li, Haoyu Wang, Hongming Zhang, Xiaodong Yu, Alexander Dong, Zhenhailong Wang, Yi~Fung, Piyush Mishra, Qing Lyu, D{\'\i}dac Sur{\'\i}s, Brian Chen, Susan~Windisch Brown, Martha Palmer, Chris Callison-Burch, Carl Vondrick, Jiawei Han, Dan Roth, Shih-Fu Chang, and Heng Ji. 2021.
\newblock \href {https://doi.org/10.18653/v1/2021.naacl-demos.16} {{RESIN}: A dockerized schema-guided cross-document cross-lingual cross-media information extraction and event tracking system}.
\newblock In \emph{Proceedings of the 2021 Conference of the North American Chapter of the Association for Computational Linguistics: Human Language Technologies: Demonstrations}, pages 133--143, Online. Association for Computational Linguistics.

\bibitem[{Wu et~al.(2020)Wu, Wang, Yuan, Wu, and Li}]{corefqa}
Wei Wu, Fei Wang, Arianna Yuan, Fei Wu, and Jiwei Li. 2020.
\newblock \href {https://doi.org/10.18653/v1/2020.acl-main.622} {{C}oref{QA}: Coreference resolution as query-based span prediction}.
\newblock In \emph{Proceedings of the 58th Annual Meeting of the Association for Computational Linguistics}, pages 6953--6963, Online. Association for Computational Linguistics.

\bibitem[{Yu et~al.(2022)Yu, Yin, and Roth}]{pairwise}
Xiaodong Yu, Wenpeng Yin, and Dan Roth. 2022.
\newblock Pairwise representation learning for event coreference.
\newblock In \emph{Proceedings of the 11th Joint Conference on Lexical and Computational Semantics}, Seattle, Washington. Association for Computational Linguistics.

\bibitem[{Zhang et~al.(2020)Zhang, Xing, Kong, Li, and Zhou}]{zhang}
Longyin Zhang, Yuqing Xing, Fang Kong, Peifeng Li, and Guodong Zhou. 2020.
\newblock \href {https://doi.org/10.18653/v1/2020.acl-main.569} {A top-down neural architecture towards text-level parsing of discourse rhetorical structure}.
\newblock In \emph{Proceedings of the 58th Annual Meeting of the Association for Computational Linguistics}, pages 6386--6395, Online. Association for Computational Linguistics.

\end{thebibliography}

\clearpage

\section{Appendices}

\appendix
\section{Appendix: The details of WEC-Zh dataset}
\label{app:A}
\subsection{Event Mention Details}

To investigate the distribution of coreference clusters, we counted the number of clusters corresponding to different mention scopes, as depicted in Table ~\ref{table.2}. 

We also conducted headword statistics for event mentions and performed Named Entity Recognition (NER) tagging on the headwords, as shown in Table ~\ref{table.8}. Additionally, we carried out Part-of-Speech (POS) tagging on the headwords, and the results are presented in Table ~\ref{table.9}.

To illustrate the diversity of events in the dataset, we conducted a statistical analysis of the unique mention count and unique head count across three partitions, as presented in Table ~\ref{table.10}.

\begin{table}[ht]
\centering
\renewcommand{\arraystretch}{1.25}
\begin{tabular}{lccc}
\hline
 & {Train} & {Dev} & {Test} \\
\hline
{>=2} & {2,745} & {213} & {248}\\
{>=5} & {1,477} & {138} & {144}\\
{>=10} & {793} & {64} & {66}\\
{>=20} & {459} & {0} & {0}\\
{>=50} & {163} & {0} & {0}\\
\hline
\end{tabular}
\caption{\label{table.2}
Statistics of clusters containing different mention numbers (>=2 indicates that the cluster contains no less than 2 mentions).}
\end{table}
\begin{table}[ht]
\centering
  \resizebox{0.4\textwidth}{!}{
\renewcommand{\arraystretch}{1.25}
\begin{tabular}{lccc}
\hline
 & {Train} & {Dev} & {Test} \\
\hline
{UNK} & {23,749} & {1,021} & {1,067} \\
{EVENT} & {22,951} & {431} & {481}\\
{DATE} & {787} & {7} & {28}\\
{FAC} & {714} & {18} & {5}\\
{WORK\_OF\_ART} & {491} & {31} & {23}\\
{LOC} & {432} & {9} & {24}\\
{GPE} & {431} & {3} & {9}\\
{ORG} & {129} & {15} & {14}\\
{CARDINAL} & {53} & {1} & {5}\\
{ORDINAL} & {48} & {0} & {0}\\
{PERSON} & {40} & {1} & {6}\\
{LAW} & {21} & {0} & {4}\\
{QUANTITY} & {10} & {1} & {1}\\
{NORP} & {4} & {0} & {0}\\
{TIME} & {1} & {0} & {0}\\
\hline
\end{tabular}}
\caption{\label{table.8}
Statistics of entity types for headword.}
\end{table}

\begin{table}[t]
\centering
{\fontsize{6}{8}\selectfont
  \resizebox{0.4\textwidth}{!}{
\renewcommand{\arraystretch}{1.25}
\begin{tabular}{lccc}
\hline
 & {Train} & {Dev} & {Test} \\
\hline
{NOUN} & {42,522} & {1,322} & {1,412} \\
{VERB} & {4,470} & {191} & {209}\\
{PROPN} & {2,337} & {15} & {24}\\
{PUNCT} & {173} & {6} & {14}\\
{ADV} & {19} & {0} & {7}\\
{NUM} & {315} & {4} & {1}\\
{DET} & {13} & {0} & {0}\\
{PRON} & {4} & {0} & {0}\\
{OTHERS} & {3} & {0} & {0}\\
{ADJ} & {2} & {0} & {0}\\
{PART} & {2} & {0} & {0}\\
{ADP} & {1} & {0} & {0}\\
\hline
\end{tabular}
}}
\caption{\label{table.9} 
Statistics of POS tags for headword.}
\end{table}

\begin{table}[t]
\centering
\renewcommand{\arraystretch}{1.25}
\begin{tabular}{lccc}
\hline
 & {Train} & {Dev} & {Test} \\
\hline
{unique-mentions}  & {9,402} & {563} & {616} \\
{unique-heads} & {2,119} & {230} & {239}\\

\hline
\end{tabular}
\caption{\label{table.10}
Statistics of unique mentions and heads.}
\end{table}

\begin{table}[h]
\centering
\renewcommand{\arraystretch}{1.25}
\begin{tabular}{lcc}
\hline
 & {WEC-Eng} & {WEC-Zh}  \\
\hline
{Our-model} & {65.0} & {66.9}\\
{Background} & {64.7} & {66.8} \\
{Elaboration} & {64.8} & {65.5} \\
{Explanation} & {64.6} & {65.6}\\
{Enablement} & {64.8} & {66.7}\\
{Cause} & {64.7} & {66.9}\\
{Contrast} & {64.9} & {66.1}\\
{Comparison} & {64.7} & {66.4}\\
{Evaluation} & {64.6} & {66.8}\\
{Summary} & {64.9} & {66.2}\\
{Temporal} & {64.8} & {66.7}\\
\hline
\end{tabular}
\caption{The CoNLL F1 scores of rhetorical ablation experiments on WEC-Eng and WEC-Zh.\label{table.retorical ablation}
}
\end{table}

\subsection{Sample Demonstration}
We have provided some Chinese dataset samples in Table \ref{table.zh_case}.

\begin{CJK}{UTF8}{gbsn}
\begin{table*}[ht]
    \centering
    \renewcommand{\arraystretch}{2.0}
    \resizebox{\textwidth}{!}{
    \begin{tabular}{l| c c c c c}
    \toprule
    \multicolumn{1}{c}{\textbf{Event mention and context}} & \multicolumn{1}{c}{\textbf{G}}  & \multicolumn{1}{c}{\textbf{W}} &\multicolumn{1}{c}{\textbf{O}}  & \multicolumn{1}{c}{\textbf{-R}} & \multicolumn{1}{c}{\textbf{-L}} \\
    \midrule
        \makecell[l]{
        \\
   公元290年，晋武帝去世，晋惠帝司马衷继位，司马衷“不慧”，至皇后贾南风专政，后八王之乱爆发，晋惠帝被毒杀，\\
   晋怀帝司马炽继位，八王之乱致西晋国力大损，各游牧民 族纷纷起兵入侵，并建立政权，引发\emph{\textcolor{myblue}{五胡乱华}}，公元311年\\
   汉赵皇帝刘聪攻入洛阳，晋怀帝被掳后被杀，晋愍帝司马邺在长安继位，不久长安亦被汉赵攻陷，西晋亡。\\ 
   In 290 AD, Emperor Wu of Jin passed away, and Emperor Hui of Jin, Sima Zhong, succeeded to the throne. Sima Zhong\\
   was “not wise” and ruled by Empress Jia Nanfeng. Later, the Eight Kings Rebellion broke out, and Emperor Huiof Jin\\
   was poisoned, Emperor Huai of Jin, Sima Chi, succeeded to the throne, and the Rebellion of the Eight Kings caused a \\
   great loss of national strength in the Western Jin Dynasty. Various nomadic ethnic groups rose up to invade and establish\\
   political power, triggering \emph{\textcolor{myblue}{Upheaval of the Five Barbarians}}. In 311 AD Emperor Liu Cong of Han and Zhao invaded\\
   Luoyang, Emperor Huai of Jin was captured and killed, and Emperor Min of Jin, Sima Ye, succeeded to the throne in\\
   Chang an. Shortly after, Chang an was also captured by Han and Zhao, and the Western Jin Dynasty perished.
   \\
   \\
   \\
   西晋末年，中原发生八王之乱，接着\emph{\textcolor{myblue}{“匈奴、鲜卑、羯、羌、氐五个部落建立非汉族政权”}}，史称“五胡乱华”，为避\\
   战乱，大量人口举族渡江南迁，东晋在江南各地设侨州、侨郡，安置流民。京口作为重要渡口，此侨置南徐州、南兖州\\
   以及南东海、南琅琊、南兰陵、南濮阳等18郡，县治多至60多个，移民数量远超土著。\\
   In the late years of the Western Jin Dynasty, the Eight Kings Rebellion occurred in the Central Plains. Subsequently, \\
   \emph{\textcolor{myblue}{five tribes of the Hu people, including the Xiongnu, Xianbei, Jie, Qiang, and Di tribes, established non-Han political power}},\\
   known as the “Upheaval of the Five Barbarians” in history. To avoid war, a large number of people crossed the\\
   Yangtze River and migrated south. The Eastern Jin Dynasty established overseas Chinese prefectures and counties in\\
   various parts of the Jiangnan region to resettle refugees. As an important ferry, Jingkou is home to 18 counties including \\
   Nanyanzhou, and Nandonghai, Nanlangya, Nanlanling, and Nanpuyang. There are as many as 60 county governments, \\
   and the number of immigrants far exceeds that of the indigenous people.
   \\
   \\
        }
        & \multirow{1}{*}{\centering}{\textbf{\textcolor{greenrgb}{\cmark}}} 
        & \multirow{1}{*}{\textbf{\textcolor{red}{\xmark}}}
        & \multirow{1}{*}{\textbf{\textcolor{greenrgb}{\cmark}}}
        & \multirow{1}{*}{\textbf{\textcolor{greenrgb}{\cmark}}}
        & \multirow{1}{*}{\textbf{\textcolor{red}{\xmark}}}
        \\
    \midrule
        \makecell[l]{
        \\
   1996年7月，葛菲/顾俊的组合代表中国参加在美国亚特兰大举行的\emph{\textcolor{myblue}{奥运会羽毛球比赛}}女子双打项目；她们在晋级过\\
   程中所向披靡，四场赛事仅仅失了39分。决赛中，葛/顾面对当时世界排名第一的韩国组合吉永雅/张惠玉，结果葛/顾\\
   只花了36分钟，便以2比0（15-5、15-5）轻松击败对手，夺得中国首面的羽毛球奥运金牌。\\
   In July 1996, the combination of Ge Fei and Gu Jun represented China in the women's doubles event of the\\
   \emph{\textcolor{myblue}{ Olympic badminton competition}} held in Atlanta, USA; They were unbeatable in the promotion process, losing only 39\\
   points in four matches. In the final, Ge/Gu faced the then-world number one South Korean Ji Yongya/Zhang Huiyu, but\\
   in just 36 minutes, Ge/Gu easily defeated their opponents 2-0 (15-5, 15-5) to win China's first badminton Olympic \\
   gold medal.
   \\
    \\
    \\
  奥运会后，葛菲/顾俊接连赢得1997年世界羽毛球锦标赛冠军、\emph{\textcolor{myblue}{1998年亚洲运动会羽毛球比赛}}金牌和1999年世界羽\\
 毛球锦标赛冠军等等；更在四年半间以无敌姿态取得全胜佳绩，连胜纪录长达100场左右。\\
 After the Olympics, Ge Fei/Gu Jun successively won the championship of the 1997 World Badminton Championships, \\
 the gold medal of \emph{\textcolor{myblue}{the 1998 Asian Games Badminton}} Championships, and the championship of the 1999 World \\
 Badminton Championships; In four and a half years, he achieved a total victory with an invincible posture, with a winning\\
 streak of about 100 games.\\
 \\
        }
        & \multirow{1}{*}{\centering}{\textbf{\textcolor{red}{\xmark}}} 
        & \multirow{1}{*}{\textbf{\textcolor{red}{\xmark}}}
        & \multirow{1}{*}{\textbf{\textcolor{greenrgb}{\cmark}}}
        & \multirow{1}{*}{\textbf{\textcolor{red}{\xmark}}}
        & \multirow{1}{*}{\textbf{\textcolor{red}{\xmark}}}
        \\
    
    \midrule
        \makecell[l]{
        \\
    辽宁红沿河核电厂位于大连市瓦房店的温坨子村，厂区三面环海，一面与陆地接壤。复州河在厂址以南约20公里处入海。\\
    厂址区域地处华北地震区，绝大部分位于郯庐断裂带的北半部，地震活动，历史上最大地震为\emph{\textcolor{myblue}{1975年海城7.3级地震}}，\\
    震中距厂址约143公里。\\
    The Liaoning Hongyanhe Nuclear Power Plant is located in Wentuozi Village, Wafangdian, Dalian City. The factory area\\
    is surrounded by the sea on three sides and borders land on one side. The Fuzhou River flows into the sea about 20\\
    kilometers south of the factory site. The factory site area is located in the North China earthquake zone, with the majority \\
    located in the northern half of the Tanlu fault zone. It is seismically active, with the largest earthquake in history being \\
    \emph{\textcolor{myblue}{the 1975 Haicheng 7.3 magnitude earthquake}}, with an epicenter approximately 143 kilometers from the factory site.
    \\
    \\
    \\
    1975年2月4日凌晨，辽宁人民广播电台直接播出地震预报。”辽宁省省委做出指示：“从当天晚上起，辽南地区海城、\\
    营口两县，所有人员都不要住在室内，生产队的大牲口、农业机械都要拉到室外。各级干部、党员、民兵全部下去，挨家\\
    挨户动员老百姓。在生产队和城镇的居民区，用大喇叭广播动员 群众。”当日19点36分，海城地区爆发\emph{\textcolor{myblue}{7.3级强烈地震}}，\\
    由于发布消息及时，海城伤亡人数为：伤一万余人，死一千三百余人，损失较小，挽救了10多万人的生命。\\
    On the early morning of February 4, 1975, Liaoning People's Broadcasting Station directly broadcasted earthquake \\
    predictions The Liaoning Provincial Party Committee has issued instructions: "Starting from the evening of that day, all \\
    personnel in Haicheng and Yingkou counties in the southern region of Liaoning should not live indoors, and the large \\
    livestock and agricultural machinery of the production team should be pulled outdoors. Cadres, party members, and \\
    militia at all levels should go down and mobilize the people door-to-door. In the residential areas of the production team \\
    and the town, loudspeakers should be used to mobilize the people." At 19:36 on the same day, a \\
    \emph{\textcolor{myblue}{strong earthquake measuring 7.3}} occurred in Haicheng area, Due to the timely release of information, the number of \\
    casualties in Haicheng was: over 10,000 injured and over 1,300 dead, with relatively small losses and saving over \\
    100,000 lives.\\
    \\
        }
        & \multirow{1}{*}{\centering}{\textbf{\textcolor{greenrgb}{\cmark}}} 
        & \multirow{1}{*}{\textbf{\textcolor{red}{\xmark}}}
        & \multirow{1}{*}{\textbf{\textcolor{greenrgb}{\cmark}}}
        & \multirow{1}{*}{\textbf{\textcolor{greenrgb}{\cmark}}}
        & \multirow{1}{*}{\textbf{\textcolor{red}{\xmark}}} 
        \\

    \bottomrule
    \end{tabular}
    }
    \caption{Case studies on some Chinese samples of the WEC-Zh dataset. \textbf{\textcolor{greenrgb}{\cmark}}indicates co-reference, \textbf{\textcolor{red}{\xmark}} indicates non co-reference,
    ``G" indicates Gold, ``O" indicates ``Our-model", 
    ``- R" indicates the removal of the RST module, ``- L" signifies the removal of lexical chains.}
    \label{table.zh_case}
\end{table*}
\end{CJK}

\section{Appendix: Error analysis}\label{app:B}
\subsection{Error Case Studies}\label{app:B.Case Studies}

To further explore the performance of our method on different specific examples and to intuitively demonstrate the assistance of RST and lexical chains in analyzing challenging event coreference cases, we selected several categories of examples from the error cases of the WEC-Eng model for experimentation with our model. 
The selected examples include: \ding{172} Highly similar mentions, \ding{173} Mentions with different semantic expressions but the same meaning, \ding{174} Mentions with remote dependencies on event parameters. The experimental results are shown in Table ~\ref{table.en_case}.

The results reveal that for highly similar mentions like \emph{\textcolor{myblue}{Eurovision Song Contest 1980}} and \emph{\textcolor{myblue}{Eurovision Song Contest 1969}}, lexical chains determine the correctness of coreference judgments. This is because the lexical chains, which include time-related information such as (1980-1969) and location information like (The Hague-Madrid), provide direct evidence for non-coreference of the events. 
Additionally, the structural information related to ``recorded by" and ``written by" also provides some corroborative information but does not play a decisive role.

For mentions with different expressions but the same semantics, such as \emph{\textcolor{myblue}{Assassination of Robert F. Kennedy}} and \emph{\textcolor{myblue}{shooting}}, there is no apparent rhetorical logical relationship between these events. 
However, there are abundant time, location, and proper noun expressions, so lexical chains play a crucial role in coreference judgment for these events. The model can construct the following lexical chains: (Senator Robert F.Kennedy -- Kennedy -- Robert F.Kennedy), (June 5, 1968 -- June 5, 1968), (the kitchen of the Ambassador Hotel in Los Angeles -- The Ambassador Hotel in Los Angeles), (1968 Democratic presidential primary in California -- Democratic 1968 California U.S. presidential primary). 
Although the event descriptions are different, there is a significant amount of repeated content related to event mentions, making it easier for the model to determine coreference for these events.

For events with remote dependencies in mentions, such as \emph{\textcolor{myblue}{42nd American Music Awards ceremony}} and \emph{\textcolor{myblue}{American Music Awards}}, there is a clear temporal evolution of the event parameters, for example, (November 24, 2014 $\rightarrow$ One day before) -- (in November 16, 2014 $\rightarrow$ On November 23 $\rightarrow$ On November 28). 
Therefore, RST can accurately capture the temporal evolution relationship between events. 
Combined with lexical chains information like (song -- music -- band), it enables the model to establish a connection between the two events, providing support for the model's judgment.

\begin{table*}[ht]
    \centering
    \resizebox{\textwidth}{!}{
    \begin{tabular}{l| c c c c c}
    \toprule
    \multicolumn{1}{c}{\textbf{Event mention and context}} & \multicolumn{1}{c}{\textbf{G}}  & \multicolumn{1}{c}{\textbf{W}} &\multicolumn{1}{c}{\textbf{O}}  & \multicolumn{1}{c}{\textbf{-R}} & \multicolumn{1}{c}{\textbf{-L}} \\
    \midrule
        \makecell[l]{
    Kennedy (known as Jean Kennedy Smith following her 1956 marriage to Stephen Edward Smith) was intricately \\
    involved with the political career of her older brother John. She worked on his 1946 Congressional campaign, \\
    his 1952 Senate campaign, and ultimately his presidential campaign in 1960. She and her siblings helped \\
    Kennedy knock on doors in primary states like Texas and Wisconsin and on the campaign trail played the role \\
    of sister more than volunteer, citing her parent's family lesson of working together for something. Smith and  \\
    her husband was present at The Ambassador Hotel in Los Angeles on June 5, 1968, during the \\
    \emph{\textcolor{myblue}{Assassination of Robert F. Kennedy}} after he had won the Democratic 1968 California  U.S. presidential primary.\\ 
    \\
    Bobby is a 2006 American drama film written and directed by Emilio Estevez, and starring an ensemble cast \\
    featuring Harry Belafonte, Joy Bryant, Nick Cannon, Laurence Fishburne, Spencer Garrett, Helen Hunt, \\
    Anthony Hopkins, Ashton Kutcher, Shia LaBeouf, Lindsay Lohan, William H.Macy, Demi Moore, Martin Sheen, \\Christian Slater, Sharon Stone, Freddy Rodriguez, Heather Graham, Elijah Wood and Estevez himself. The \\ screenplay is a fictionalized account of the hours leading up to the June 5, 1968 \emph{\textcolor{myblue}{shooting}} of U.S. Senator \\
    Robert F. Kennedy in the kitchen of the Ambassador Hotel in Los Angeles following his win of the 1968 \\Democratic presidential primary in California.
        }
        & \multirow{1}{*}{\centering}{\textbf{\textcolor{greenrgb}{\cmark}}} 
        & \multirow{1}{*}{\textbf{\textcolor{red}{\xmark}}}
        & \multirow{1}{*}{\textbf{\textcolor{greenrgb}{\cmark}}}
        & \multirow{1}{*}{\textbf{\textcolor{greenrgb}{\cmark}}}
        & \multirow{1}{*}{\textbf{\textcolor{red}{\xmark}}}
        \\
    \midrule
        \makecell[l]{
   The song was released on November 24, 2014 on iTunes. The song was also released on Spotify, make it \\
   available to stream online. One day before the release date, Lil Wayne premiered the single at\\ 
     \emph{\textcolor{myblue}{42nd American Music Awards ceremony}} with Christina Milian.
   \\
    \\
   The first ``Smile" performance was in Family Channel special Family Day, in Canada, on November 16, 2014.\\
   On November 23, the band performed at the \emph{\textcolor{myblue}{American Music Awards}}. On November 28, they performed in \\Good Morning America.
        }
        & \multirow{1}{*}{\centering}{\textbf{\textcolor{greenrgb}{\cmark}}} 
        & \multirow{1}{*}{\textbf{\textcolor{red}{\xmark}}}
        & \multirow{1}{*}{\textbf{\textcolor{greenrgb}{\cmark}}}
        & \multirow{1}{*}{\textbf{\textcolor{red}{\xmark}}}
        & \multirow{1}{*}{\textbf{\textcolor{red}{\xmark}}}
        \\
    
    \midrule
        \makecell[l]{
    ``Quédate esta noche" is a song recorded by Spanish group Trigo Limpio. The song was written by José \\
    Antonio Martín. It is best known as the Spanish entry at the \emph{\textcolor{myblue}{Eurovision Song Contest 1980}}, in The Hague.\\
    \\
    ``Boom Bang-a-Bang" is a song recorded by British singer Lulu. The song was written by Alan Moorhouse and\\ Peter Warne. It is best known as the British winning entry at the \emph{\textcolor{myblue}{Eurovision Song Contest 1969}}, held in Madrid.
        }
        & \multirow{1}{*}{\centering}{\textbf{\textcolor{red}{\xmark}}} 
        & \multirow{1}{*}{\textbf{\textcolor{red}{\xmark}}}
        & \multirow{1}{*}{\textbf{\textcolor{greenrgb}{\cmark}}}
        & \multirow{1}{*}{\textbf{\textcolor{greenrgb}{\cmark}}}
        & \multirow{1}{*}{\textbf{\textcolor{red}{\xmark}}} 
        \\

    \midrule
        \makecell[l]{
    Along with von Einem, he is considered to be the best suspect for the Beaumont children abduction as he bore \\
    a striking similarity to an identikit picture of the suspect for both the Beaumont children and Adelaide Oval cases. 
    \\A search for a connection to the Beaumonts was unsuccessful as no employment records existed that could \\
    shed light on his movements at the time. Some of the records were believed lost in the \emph{\textcolor{myblue}{1974 Brisbane flood}} \\
    and it is also possible that Brown, who had unrestricted access to government buildings, may have deleted \\
    his files.
    \\
    \\
    During World War II he served in the Royal Australian Air Force as a navigator and intelligence officer. \\
    Returning to Queensland, he farmed sugarcane and pineapples and joined the Liberal Party. In 1963 he was \\
    elected to the Queensland Legislative Assembly as the member for Mount Coot-tha. On 3 October 1975, he was \\
    awarded the Queen's Gallantry Medal for his efforts to rescue a soldier during the \emph{\textcolor{myblue}{flooding in Brisbane}} \\ 
    the previous year. In 1975 he was appointed to the front bench as Minister for Survey, Valuation, Urban and \\
    Regional Affairs, with a further promotion to Attorney-General and Minister for Justice in 1976.
        }
        & \multirow{1}{*}{\centering}{\textbf{\textcolor{greenrgb}{\cmark}}} 
        & \multirow{1}{*}{\textbf{\textcolor{red}{\xmark}}}
        & \multirow{1}{*}{\textbf{\textcolor{greenrgb}{\cmark}}}
        & \multirow{1}{*}{\textbf{\textcolor{greenrgb}{\cmark}}}
        & \multirow{1}{*}{\textbf{\textcolor{greenrgb}{\cmark}}} 
        \\

        \midrule
        \makecell[l]{
   In September 2017, New York's 25th Assembly District Representative Nily Rozic, a Democrat, suggested \\
   renaming the park in honor of Heather Heyer, who died in the \emph{\textcolor{myblue}{Charlottesville car attack}} protesting the \\
   Unite the Right rally in Charlottesville, Virginia. No renaming of the park has been undertaken and the proposal \\
   was largely ignored. Rozic, along with New York State Senator Brad Hoylman, reintroduced legislation \\
   to rename the park in 2018 but it did not make it out of committee. They again introduced legislation in 2019 \\
   to rename the park.
    \\
    \\
   After covering a Ku Klux Klan rally in Charlottesville, Virginia, she received permission to meet with Jeff \\ Schoep, the leader of the National Socialist Movement. Afterwards, she receives permission to film the group \\
   at the Unite the Right rally in Charlottesville, where the group gets into an altercation with and are \\
   pepper-sprayed by Antifa counter-protestors. After \emph{\textcolor{myblue}{the death of Heather Heyer}} and President Donald Trump's\\
   controversial remarks on the rally, Schoep takes Deeyah to the urban decay in Detroit and explains that he \\moved the organization's headquarters to the city to take advantage of its economic decline for recruiting.
        }
        & \multirow{1}{*}{\centering}{\textbf{\textcolor{greenrgb}{\cmark}}} 
        & \multirow{1}{*}{\textbf{\textcolor{red}{\xmark}}}
        & \multirow{1}{*}{\textbf{\textcolor{red}{\xmark}}}
        & \multirow{1}{*}{\textbf{\textcolor{red}{\xmark}}}
        & \multirow{1}{*}{\textbf{\textcolor{red}{\xmark}}} 
        \\
        
    \bottomrule
    \end{tabular}
    }
    \caption{Case studies on some samples of the WEC-Eng dataset. \textbf{\textcolor{greenrgb}{\cmark}}indicates co-reference, \textbf{\textcolor{red}{\xmark}} indicates non co-reference,
    ``G" indicates Gold, ``O" indicates ``Our-model", 
    ``- R" indicates the removal of the RST module, ``- L" signifies the removal of lexical chains.}
    \label{table.en_case}
\end{table*}

\section{Appendix: Other experiments}\label{app:C}

\subsection{Ablation Experiments on Different Rhetorical Relations}
To more deeply explore the impact of different rhetorical relations on the effectiveness of coreference models, we conducted ablation experiments on the following rhetorical relations: Background, Elaboration, Attribution, Same-Unit, Joint, Explanation, Enablement, Cause, Topic-Comment, Contrast, Condition, Comparison, Evaluation, Manner-Means, Summary, Temporal, Topic-Change, Textual-Organization.The experi-
mental results are shown in Table \ref{table.retorical ablation}.

The specific approach was to remove the edges corresponding to these rhetorical relations in the constructed graph and add self-loop edges to the corresponding nodes. The experimental results show that for the WEC-Eng test set, the removal of Explanation, Evaluation, Background, and Cause led to the most significant decrease in model performance, with conll f1 scores dropping by 0.4, 0.4, 0.3, and 0.3, respectively. The conll f1 decreases for Temporal, Elaboration, and Enablement were around 0.2, while the remaining rhetorical relations had a minimal impact on the decrease in model performance. 
For the WEC-Zh test set, the rhetorical relations of Elaboration and Explanation have the most significant impact on the model, with conll f1 scores decreasing by 1.4 and 1.3 respectively. Additionally, the conll f1 scores for Contrast, Summary, and Comparison decreased by 0.8, 0.7, and 0.5 respectively. The impact of the remaining rhetorical relations on the model was relatively minor. This indicates that Elaboration and Explanation play critical roles in the performance of the model on the WEC-Zh test set, highlighting their importance in understanding and modeling discourse. 
The results also suggest areas for potential improvement in the model, particularly in handling these specific rhetorical relations more effectively.

\end{document}